\documentclass{article}

\usepackage[eandd, final]{neurips_2026}

\usepackage[utf8]{inputenc} 
\usepackage[T1]{fontenc}    
\usepackage{hyperref}       
\usepackage{url}            
\usepackage{booktabs}       
\usepackage{amsfonts}       
\usepackage{nicefrac}       
\usepackage{microtype}      
\usepackage{xcolor}         
\usepackage{wrapfig}
\usepackage{float}

\usepackage{amsmath,amssymb,amsthm}
\usepackage{booktabs}
\usepackage{multirow}
\usepackage{graphicx}
\usepackage[table]{xcolor}
\usepackage{colortbl}
\usepackage{array}
\usepackage{caption}
\usepackage{subcaption}
\usepackage{tikz}
\usepackage{pgfplots}
\usepackage{enumitem}
\usepackage{pifont}
\usepackage{float}
\usepackage{tikz}
\usepackage{xcolor}
\definecolor{stagebox}{HTML}{F0A45B}
\usepackage[subtle]{savetrees}
\pgfplotsset{compat=1.18}

\newtheorem{definition}{Definition}
\newtheorem{proposition}{Proposition}

\definecolor{condA}{HTML}{FCC495}       
\definecolor{condBauto}{HTML}{E0E2B5}   
\definecolor{condBperc}{HTML}{D1ECF1}   
\definecolor{condBfull}{HTML}{9DD1E2}   
\definecolor{condC}{HTML}{D5C6E0}       
\definecolor{stageone}{HTML}{FCC495}    
\definecolor{stagetwo}{HTML}{D1ECF1}    
\definecolor{stagethree}{HTML}{F0A45C}  
\definecolor{condAbest}{HTML}{F0A45C}   
\definecolor{condCbest}{HTML}{B8A4CC}   
\definecolor{rankfirst}{HTML}{9DD1E2}   
\definecolor{ranksecond}{HTML}{D1ECF1}  
\definecolor{rankthird}{HTML}{E0E2B5}   
\definecolor{rescue}{HTML}{D1ECF1}      
\definecolor{hurt}{HTML}{FCC495}        
\definecolor{condBvlm}{HTML}{C8E6C9}    
\definecolor{condBstrip}{HTML}{FFE0B2}   
\definecolor{deltaup}{HTML}{00897B}     
\definecolor{deltadn}{HTML}{C62828}     
\newcommand{\up}[1]{{\scriptsize\color{deltaup}(#1)}}
\newcommand{\dn}[1]{{\scriptsize\color{deltadn}(#1)}}

\newcommand{\chipA}{\tikz[baseline=-0.7ex]{\node[fill=condA, rounded corners=2pt, minimum width=7pt, minimum height=7pt, inner sep=0pt]{};}}
\newcommand{\chipBa}{\tikz[baseline=-0.7ex]{\node[fill=condBauto, rounded corners=2pt, minimum width=7pt, minimum height=7pt, inner sep=0pt]{};}}
\newcommand{\chipBp}{\tikz[baseline=-0.7ex]{\node[fill=condBperc, rounded corners=2pt, minimum width=7pt, minimum height=7pt, inner sep=0pt]{};}}
\newcommand{\chipBf}{\tikz[baseline=-0.7ex]{\node[fill=condBfull, rounded corners=2pt, minimum width=7pt, minimum height=7pt, inner sep=0pt]{};}}
\newcommand{\chipC}{\tikz[baseline=-0.7ex]{\node[fill=condC, rounded corners=2pt, minimum width=7pt, minimum height=7pt, inner sep=0pt]{};}}
\newcommand{\chipBv}{\tikz[baseline=-0.7ex]{\node[fill=condBvlm, rounded corners=2pt, minimum width=7pt, minimum height=7pt, inner sep=0pt]{};}}
\definecolor{condAvar}{HTML}{FDE6D2}   
\newcommand{\chipAv}{\tikz[baseline=-0.7ex]{\node[fill=condAvar, draw=condA, rounded corners=2pt, minimum width=7pt, minimum height=7pt, inner sep=0pt]{};}}
\newcommand{\chipBs}{\tikz[baseline=-0.7ex]{\node[fill=condBstrip, rounded corners=2pt, minimum width=7pt, minimum height=7pt, inner sep=0pt]{};}}
\newcommand{\legendchip}[2]{%
  \tikz[baseline=(X.base)]{%
    \node[fill=#1, rounded corners=2.5pt, inner sep=2pt,
          minimum width=14pt, minimum height=13pt](X){%
      \vphantom{$B_f$}\scriptsize #2};%
  }%
}
\newcommand{\stagepanel}[1]{%
  \begin{tikzpicture}
    \node[draw=stagebox, fill=stagebox!15, line width=2pt,
          rounded corners=8pt, inner sep=10pt] (box) {%
      \begin{minipage}{0.92\linewidth}
        \centering
        #1
      \end{minipage}};
  \end{tikzpicture}%
}

\usepackage{xspace}
\makeatletter
\DeclareRobustCommand\onedot{\futurelet\@let@token\@onedot}
\def\@onedot{\ifx\@let@token.\else.\null\fi\xspace}

\def\eg{\emph{e.g}\onedot}

 \def\vs{\emph{vs}\onedot}

\def\etal{\emph{et al}\onedot}
\makeatother

\title{Perception, Not Reasoning, Limits\\Visual Theory of Mind}

\vspace{-0.5em}
\author{%
\textbf{Mohamed~Rayan~Barhdadi}$^{1}$\quad
\textbf{Syed~Talal~Wasim}$^{2,3}$\quad
\textbf{Juergen~Gall}$^{2,3}$\\
\textbf{Erchin~Serpedin}$^{1}$\quad
\textbf{Hasan~Kurban}$^{4}$\\[4pt]
{\normalfont\normalsize $^{1}$Texas A\&M University\quad
$^{2}$University of Bonn}\\
{\normalfont\normalsize $^{3}$Lamarr Institute for Machine Learning and Artificial Intelligence}\\
{\normalfont\normalsize $^{4}$Hamad Bin Khalifa University}
}

\begin{document}

\maketitle

{\renewcommand{\thefootnote}{\fnsymbol{footnote}}%
\footnotetext{Corresponding authors: \texttt{\small rayan.barhdadi@tamu.edu} and \texttt{\small hkurban@hbku.edu.qa}.}}

\newcommand{\condlegend}{%
\noindent{\scriptsize\legendchip{condA}{A}\;\legendchip{condBauto}{$B_a$}\;\legendchip{condBperc}{$B_p$}\;\legendchip{condBfull}{$B_f$}\;\legendchip{condC}{C}}%
}

\vspace{-1.3em}

\begin{abstract}
\vspace{-0.4em}
Vision-language models pass text-based false-belief probes at near-human levels but drop by over twenty percentage points on abstract panels the moment the same problem is conveyed visually. The consensus diagnosis is conceptual, prescribing new belief-reasoning modules. We argue the dominant bottleneck is perceptual: not what the model cannot reason about, but what it cannot see. We propose a five-condition dissociation protocol that holds the reasoning task constant while varying perceptual fidelity from visual only to text only, and quantify the result with the Perceptual Contribution Ratio, the share of the visual-to-text gap a given fidelity closes. We test this on two kinds of scenes: abstract 2D panels in the Heider–Simmel tradition, and naturalistic 3D game scenes from MuMA-ToM and MindPower. The evaluation covers seven VLMs and roughly 159,000 API calls. Ground-truth scaffolds substantially close the gap in every setting, and on the abstract panels and MuMA-ToM they match or exceed the text-only reasoning. Whereas scaffolds built by automatic parsers, including frontier VLMs, fail. The most common error in the visual-only condition is \emph{reality bias}, which accounts for 60\% of false-belief failures. Models report where an object actually is, not where an absent agent believes it is. This points to a perceptual front-end that fails to track which agents were present at each moment. Under the evaluated protocol, structured visual scene-state extraction, rather than belief inference, is the dominant bottleneck for visual ToM. Project page: \href{https://kurbanintelligencelab.github.io/perception-not-reasoning/}{\nolinkurl{kurbanintelligencelab.github.io/perception-not-reasoning}}.
\end{abstract}

\vspace{-16pt}
\begin{figure}[!h]
    \centering
    \includegraphics[width=0.84\linewidth]{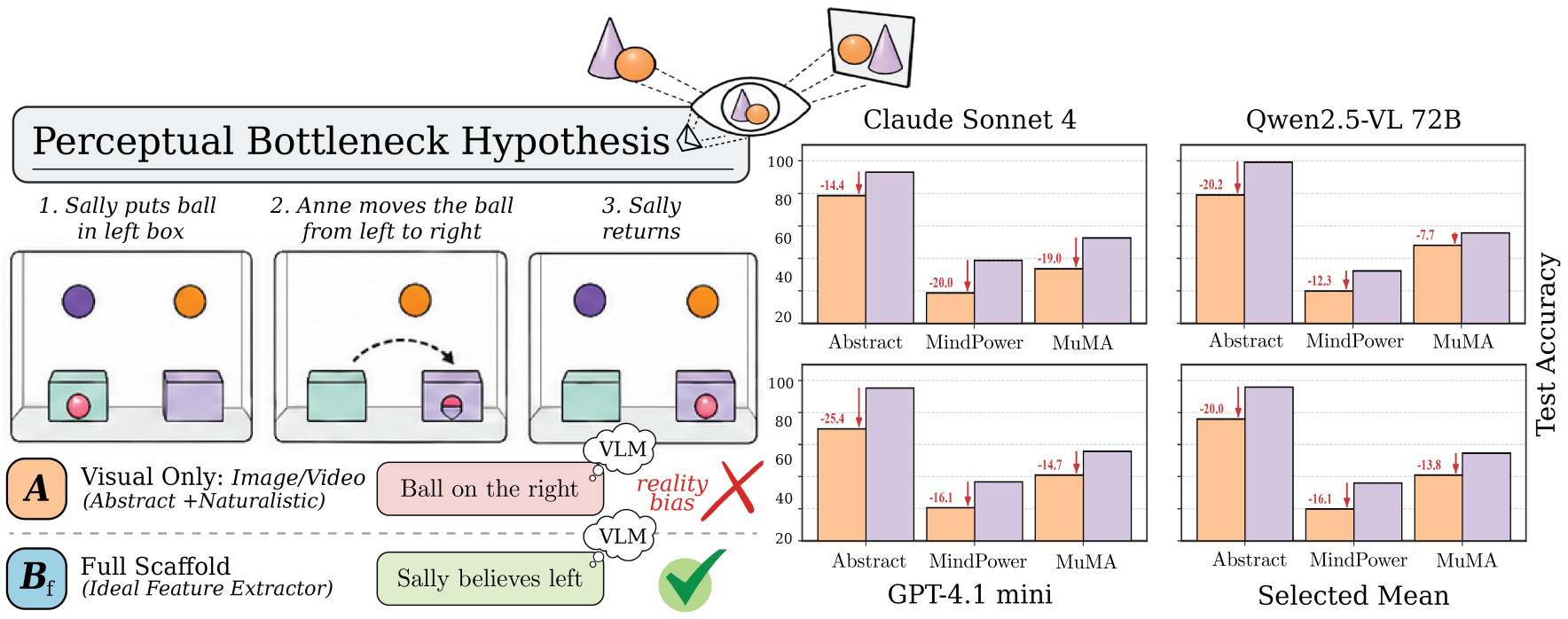}
    \caption{\textbf{Visual Theory of Mind is limited by perceptual scene-state extraction.} \textit{Left:} on a Sally-Anne probe, the VLM under visual-only input (A) reports the ball's current location (\emph{reality bias}); a ground-truth scaffold ($B_f$) recovers the absent agent's belief. \textit{Right:} per-model visual-only vs.\ text-only accuracy on abstract 2D panels, MindPower~\cite{zhang2025mindpower}, and MuMA-ToM~\cite{shi2025muma}. Ground-truth scaffolds close the gap ($\mathrm{PCR}_f \!\geq\! 1$ on abstract and MuMA-ToM); automatic parsers do not.}
    \label{fig:teaser}
\end{figure}
\section{Introduction}
\label{sec:intro}

Children come to read other minds through a sequence of perceptual achievements, not a single conceptual leap. Infants individuate objects behind occluders by three to five months~\cite{spelke1990principles}, read motion as goal-directed by six to twelve months~\cite{woodward1998infants,gergely1995taking}, and pass false-belief tasks by roughly age four~\cite{wimmer1983beliefs,baron1985does,wellman2001meta}. Each capacity requires the previous as input: a goal cannot be attributed to an entity not yet individuated, and a belief cannot be modeled for an agent whose goals cannot be parsed~\cite{carey2009origin,leslie1987pretense}. The Sally-Anne false-belief task, the canonical assay for Theory of Mind, sits at the top of this chain~\cite{baron1985does,premack1978does}.

When this task was adapted for large language models, frontier systems passed text-based probes at near-human levels~\cite{kosinski2024evaluating,strachan2024testing}. The moment the same problem is conveyed visually, performance collapses by twenty or more percentage points across every VLM tested~\cite{li2024cogdevelop2k,chen2024vlms_see}, with even larger gaps on naturalistic multi-agent video~\cite{shi2025muma,jin2024mmtom,zhang2025mindpower}. The consensus diagnosis is conceptual: VLMs are taken to lack the machinery belief attribution requires, prescribing explicit social-reasoning modules atop existing visual encoders~\cite{ullman2023large,sap2022neural,shapira2024clever}.

We argue that this diagnosis is incomplete. The bottleneck is not what the model cannot reason about; it is what it cannot see. If a VLM solves a false-belief problem at $94.5\%$ in text but only $73.9\%$ in pixels, the deficit lies in the perceptual pipeline that converts pixels into propositions, not in the reasoning that operates on them. We call this the \emph{perceptual bottleneck hypothesis}: visual ToM failures reflect breakdowns in structured perceptual-state extraction (entity identity, position, presence, absence across panels or frames), not in downstream belief reasoning. We use \emph{perception} operationally for this structured scene-state extraction, including cross-frame correspondence and event detection, and \emph{reasoning} for the mapping from scene states and access facts to belief attributions and answers (\S\ref{sec:protocol}). The hypothesis is falsifiable. If the bottleneck is perceptual, giving the model the ground-truth scene description should bring performance back toward the text-only reasoning. If the bottleneck is conceptual, no scaffold will help, however accurate it is.

To test this, we propose a five-condition dissociation protocol that holds the reasoning task constant while varying perceptual fidelity from visual only to text only with three intermediate scaffolds. We instantiate it on two stimulus regimes following prior conventions: \emph{abstract 2D panels} in the Heider-Simmel tradition~\cite{heider1944experimental}, organized along the developmental cascade, and \emph{naturalistic 3D game scenes} from MuMA-ToM~\cite{shi2025muma} and MindPower~\cite{zhang2025mindpower}. We quantify scaffold benefit with the \emph{Perceptual Contribution Ratio} (PCR), a dose-response metric taking value $1$ when a scaffold fully recovers the text-only ceiling and $0$ when it does not help.

The signature reproduces across both regimes and seven VLMs spanning three capability tiers. Ground-truth scaffolds substantially close the visual-to-text gap on every regime, matching or exceeding the text-only ceiling on abstract panels and MuMA-ToM and recovering $59\%$ of it on MindPower. Automatic parsers, including frontier VLMs, fail. Stripping the explicit visibility and absence statements from the ground-truth scaffold by pattern matching retains most of the gain, ruling out that scaffolds simply hand the model the reasoning predicate. A finer content ladder attributes the largest aggregate increment to cross-frame event structure, and length-matched corrupted scaffolds recover less than half of the correct scaffold's gain on aggregate, arguing against text volume as the explanation. The most common error in the visual-only condition is \emph{reality bias}, which accounts for 60\% of false-belief failures. Models report where an object actually is, not where an absent agent believes it is. This points to a perceptual front-end that fails to track which agents were present at each moment. When the front-end fails to recover the state of each entity in the scene, the downstream reasoning cannot make up for it. Under the evaluated protocol, structured visual scene-state extraction, rather than belief inference, is the dominant bottleneck for visual ToM.

\noindent\textbf{Contributions:}
\begin{itemize}[leftmargin=*,nosep]
    \item A five-condition dissociation protocol isolating perceptual from conceptual contributions to visual ToM failures, applied uniformly across abstract 2D and naturalistic 3D regimes.
    \item The Perceptual Contribution Ratio: a dose-response metric tracing how much of the visual-to-text gap a scaffold of given fidelity closes, with negative values flagging misleading scaffolds and values above $1$ indicating the combination exceeds the text-only ceiling.
    \item Cross-regime evidence, across seven VLMs and roughly $159{,}000$ API calls, that the visual ToM deficit is predominantly perceptual rather than conceptual, supported by a scaffold-content ladder, length-matched corruption controls, and the reality-bias error taxonomy.
\end{itemize}
\section{Related work}
\label{sec:related}

\noindent\textbf{Theory of Mind in language models:} Whether LLMs possess Theory of Mind remains contested. Kosinski~\cite{kosinski2024evaluating} reported GPT-4 passes false-belief tasks at six-year-old levels; Ullman~\cite{ullman2023large} showed that trivial paraphrases cause catastrophic failures. Strachan~\etal~\cite{strachan2024testing} found above-chance belief tracking with systematic biases absent in humans; Gandhi~\etal~\cite{gandhi2023understanding} documented uneven performance across ToM dimensions; Shapira~\etal~\cite{shapira2024clever} introduced adversarial controls that degrade LLM ToM; Hu~\etal~\cite{hu2025reevaluating} argue existing evaluations conflate linguistic competence with mentalizing; Kim~\etal~\cite{kim2023fantom} stress-tested ToM in information-asymmetric conversations. These studies are text-only. In the visual domain, Jin~\etal~\cite{jin2024mmtom} introduced MMToM-QA; Shi~\etal~\cite{shi2025muma} extended this with MuMA-ToM; Chen~\etal~\cite{chen2024vlms_see} showed VLMs fail at visual perspective-taking despite intent recognition; Zhang~\etal~\cite{zhang2025mindpower} documented a substantial human-VLM gap with MindPower. None measures whether the visual deficit is perceptual or conceptual in origin. Our protocol forces that distinction by holding the reasoning task constant and varying only the fidelity of perceptual input, applied uniformly across abstract 2D panels and naturalistic 3D game scenes.

\noindent\textbf{Cognitive benchmarks for VLMs:} CogDevelop2K~\cite{li2024cogdevelop2k} evaluated 46 multimodal LLMs on 12 cognitive sub-concepts organized by developmental milestones, finding models outperform on late-developing abstract reasoning relative to early-developing perceptual tasks. Tong~\etal~\cite{tong2024eyes} traced systematic visual shortcomings to CLIP-based encoders; MotionBench~\cite{hong2025motionbench} reported substantial gaps in fine-grained video motion understanding. These benchmarks score VLMs but treat the resulting accuracy as a single number; perceptual and inferential contributions remain entangled. Our scaffolding intervention dissociates the two and assigns the deficit to perception rather than reasoning.

\noindent\textbf{Developmental foundations:} Spelke's core knowledge theory~\cite{spelke1990principles,spelke2000core} posits domain-specific representational systems with a cascaded developmental trajectory: object individuation through spatiotemporal continuity~\cite{spelke1990principles} is a prerequisite for goal attribution~\cite{woodward1998infants,gergely1995taking}, which is a prerequisite for belief modeling~\cite{wimmer1983beliefs,baron1985does,carey2009origin}. Leslie~\cite{leslie1987pretense} formalized this as a metarepresentational requirement; Onishi and Baillargeon~\cite{onishi2005do} and Baillargeon~\etal~\cite{baillargeon2010false} demonstrated implicit false-belief sensitivity in infancy; Wellman~\etal~\cite{wellman2001meta} meta-analyzed 178 studies confirming robust developmental change. Apperly and Butterfill~\cite{apperly2009two} proposed dual-systems distinguishing efficient inflexible belief tracking from flexible effortful mentalizing; Shu~\etal~\cite{shu2021agent} operationalized core psychological reasoning in AGENT. This literature establishes the cascade as a property of human development; it does not test whether VLMs are subject to a parallel dependency. Our protocol provides that test, and the dose-response signature places the bottleneck in the front-end of the cascade.

\noindent\textbf{Model-based ToM and structured intermediates:} Several approaches assume access to structured state representations as input. ToMNet~\cite{rabinowitz2018tomnet} trains a meta-learner over agent trajectories represented as state-action sequences; BIP-ALM~\cite{jin2024mmtom} combines Bayesian inverse planning with structured scene representations to achieve strong belief inference. These systems take structured intermediates as given and evaluate the reasoning that follows. Our protocol takes the inverse view: it holds the inferential pathway constant and varies the fidelity of the structured intermediate that visual encoders attempt to produce, isolating the cost of extracting that intermediate from pixels.
\section{Benchmarking protocol}
\label{sec:protocol}

We treat visual ToM as two pipelines in series: a perceptual front-end converting pixels into propositions, and an inferential back-end mapping them and a question to an answer. A single accuracy figure under visual input cannot tell us which is failing. Our protocol forces them apart by holding the inferential task constant while parametrically varying perceptual fidelity, from no scaffold to a complete ground-truth scene description. It applies identically to abstract 2D panels and naturalistic 3D game scenes.

\noindent\textbf{Operational definition:} Perception is structured scene-state extraction: entity identity, position, and presence or absence across panels or frames, including the cross-frame correspondence and event detection that link them. Like object individuation and tracking through occlusion~\cite{spelke1990principles,baillargeon2010false}, these competences carry no mental-state content. Reasoning maps such scene states and access facts to belief attributions and answers. Cross-frame event linking is an intermediate representation; we assign it to perception by this definition rather than assume the boundary, and no scaffold below $B_f$ states a belief.

\subsection{The five-condition dissociation}
\label{sec:conditions}

\begin{figure}[t]
    \centering
    \includegraphics[width=0.86\linewidth]{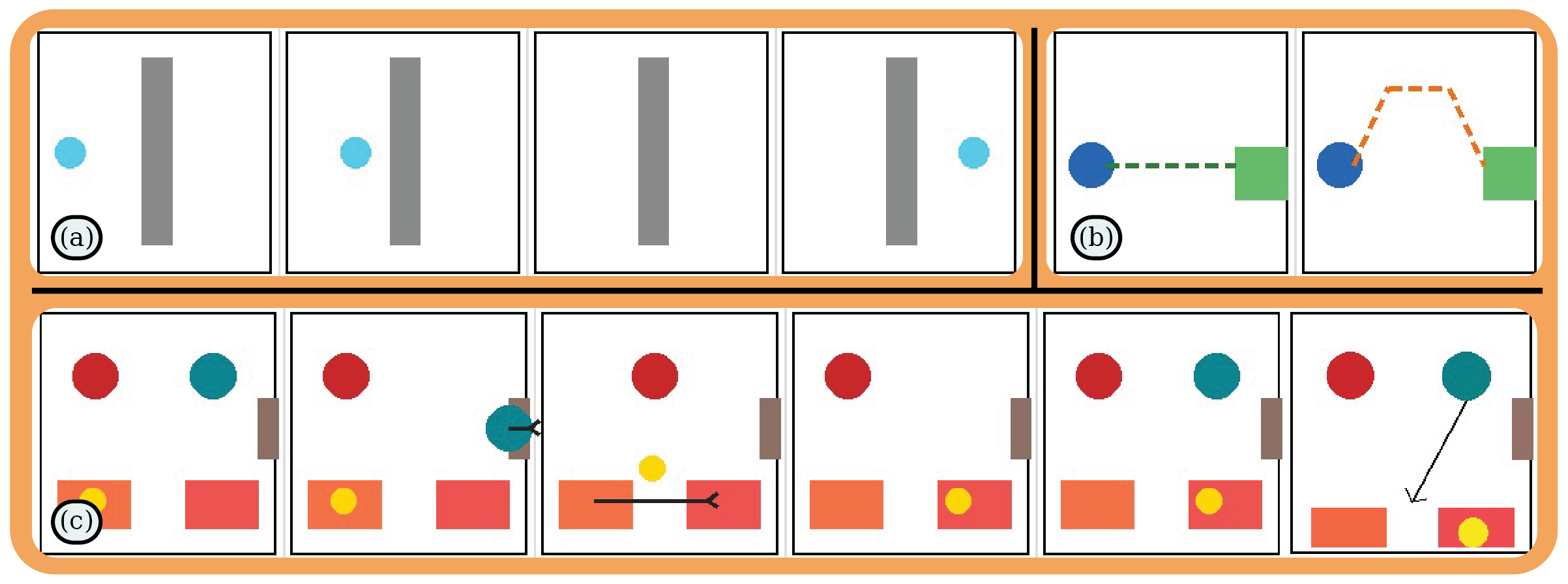}
    \caption{\textbf{Representative abstract 2D probes across the developmental cascade}, one template per stage. \textbf{(a)}~Stage~1 object tracking (\textsc{occlusion\_consistent}): a cyan circle moves rightward, becomes hidden behind a gray occluder, and reappears. \emph{Question:} ``Is the final reappearance position consistent with the earlier direction of motion, yes or no?'' (correct: yes). \textbf{(b)}~Stage~2 goal attribution (\textsc{wall\_removed}): a blue agent must reach a green goal with no obstacle present. \emph{Question:} ``With no obstacle present, which path would a goal-directed character choose, straight or curved?'' (correct: straight). \textbf{(c)}~Stage~3 false belief (\textsc{knowledge\_belief}): the teal character sees a yellow object in the left container and exits; the red character moves it to the right container; the teal character returns. \emph{Question:} where the teal character thinks the object is (correct: left). All stimuli rendered programmatically with no embedded text; full catalog in \S\ref{app:dataset_overview}.}
    \label{fig:abstract_examples}
    \vspace{-0.3cm}
\end{figure}

Each probe is evaluated under five conditions, sampling the visual-to-text fidelity continuum. A and C anchor the endpoints; the three B conditions sit between, each adding a textual scaffold of increasing fidelity to the image.

\noindent\textbf{\protect\legendchip{condA}{A}~Visual only:} Multi-panel image (abstract) or sampled video frames (naturalistic) with the question; no scene description (verbatim prompts in \S\ref{app:dataset_overview}). Measures how well the model's own front-end extracts the scene state the question requires.

\noindent\textbf{\protect\legendchip{condBauto}{$B_a$}~Image with automatic-parser scaffold:} Visual input plus a scaffold from a parser with no ground-truth access. Abstract: a hand-engineered pixel pipeline using flood-fill blob detection, color matching, and inter-panel differencing (\S\ref{app:parser}). Naturalistic: a frontier VLM (GPT-4o; additionally Claude Sonnet~4 on MindPower), prompted to produce panel-by-panel entity narrations with no ToM interpretation. Both are imperfect by construction; $B_a$ asks whether a realistic front-end recovers the scene structure belief attribution requires.

\noindent\textbf{\protect\legendchip{condBperc}{$B_p$}~Image with perceptual-grade scaffold:} Visual input plus a scaffold correct by construction but deliberately omitting visibility and absence inference: it names entities, positions, and observable events, may state that an agent witnesses an event, but never states that one agent did not observe what another did. This isolates how much scaffold benefit comes from perceptual content rather than visibility cues. Abstract: a panel-by-panel entity narration generated programmatically from the scene generator's state. MindPower: the dataset's scene description (script-derived for 427 of 456 VirtualHome scenarios, human-written otherwise), which contains observable scene state but not the belief, desire, intention, decision, or action layers used in scoring, and occasionally contains state-comparison phrasing (\eg ``the box is closed again''). Undefined for MuMA-ToM, whose annotations yield a single scaffold; it matches no visibility pattern (below), so there $B_f$ is close to perceptual-grade.

\noindent\textbf{\protect\legendchip{condBfull}{$B_f$}~Image with full ground-truth scaffold:} Visual input plus the most complete reference scaffold per regime. Abstract: $B_p$ plus explicit visibility and absence statements (``agent $X$ was not in the scene when the object was moved''); for the second-order template, the scaffold also states the resulting first-order belief. MuMA-ToM: ground-truth entity identities, locations, and actions; none of the 900 logged prompts matches the visibility patterns of \S\ref{app:visstrip} (\S\ref{app:muma_audit}). MindPower: the full human reference chain across all six layers, shared across all scored layers of a scenario; on this regime, $B_f$ functions as a full-information oracle upper bound rather than a leakage-controlled perceptual ceiling.

\noindent\textbf{\protect\legendchip{condC}{C}~Text only:} The scene conveyed as text, with no image: on abstract panels, a short narration of the same scene, including who witnessed each event (\S\ref{app:dataset_overview}); on MuMA-ToM, the $B_f$ text; on MindPower, the $B_p$ description. Measures the reasoning ceiling when no visual processing is required; $\mathrm{Acc}(C) - \mathrm{Acc}(A)$ is the visual-to-text deficit that scaffolds attempt to close.

\noindent\textbf{Why these five conditions:} Each transition adds one interpretable increment of fidelity while question and image are held fixed: A $\to B_a$ adds what a realistic automatic front-end produces; $B_a \to B_p$ makes the content correct while withholding visibility and absence; $B_p \to B_f$ adds visibility and absence; $B_f \to C$ removes the image. The endpoints are the settings prior work measures; the intermediate rungs separate front-end quality ($B_a$ \vs $B_p$) from epistemic content ($B_p$ \vs $B_f$) and from the image ($B_f$ \vs C). Accuracy therefore traces a dose-response along a single axis, the fidelity of perceptual input reaching the inferential pathway, which PCR (\S\ref{sec:pcr}) reads rung by rung; \S\ref{sec:results} refines the ladder below $B_p$.

\begin{figure}[t]
    \centering
    \includegraphics[width=0.86\linewidth]{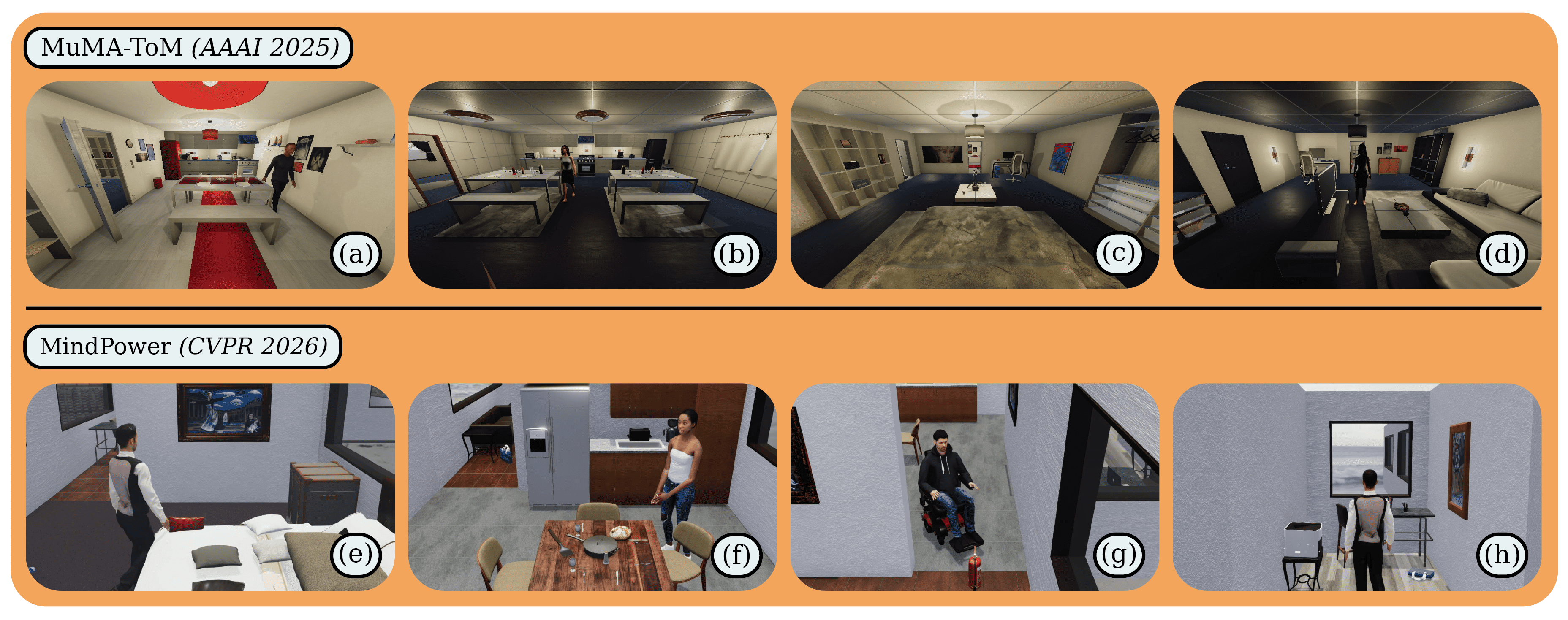}
    \caption{\textbf{Representative frames from the naturalistic multi-agent benchmarks.} \textit{Top:} MuMA-ToM~\cite{shi2025muma} household videos with human-rendered agents; questions probe \textsc{belief}, \textsc{belief\_of\_goal}, \textsc{social\_goal}. \textbf{(a)--(d)}~kitchen, kitchen-adjacent, bedroom-to-living, and dimly lit living-room layouts. \textit{Bottom:} MindPower~\cite{zhang2025mindpower} ThreeDWorld scenarios; the model acts as a robot helper inferring the human's implicit goal. \textbf{(e),(g)}~obstacle removal (pillows on a bed; a wheelchair user blocked by a fire hydrant). \textbf{(f),(h)}~missing tool (bread without a utensil; a low-battery MacBook without its cable). Evaluations use multi-frame inputs (8 frames per MuMA-ToM video, storyboards of up to 12 frames per MindPower scenario).}
    \label{fig:naturalistic_examples}
    \vspace{-0.3cm}
\end{figure}

\subsection{Stimulus regimes}
\label{sec:regimes}

The two regimes differ on every perceptual dimension. \textbf{Abstract 2D panels} present ToM structure in stripped-down form: multi-panel images of geometric shapes in the Heider-Simmel tradition~\cite{heider1944experimental}, with zero embedded text. \textbf{Naturalistic 3D game scenes} move to multi-agent household activity rendered in 3D game environments, drawing on MuMA-ToM~\cite{shi2025muma} and MindPower~\cite{zhang2025mindpower}. The two stress-test the perceptual-bottleneck account from opposite directions: abstract asks whether the bottleneck appears even when stimuli are deliberately easy to parse; naturalistic asks whether the same dose-response reproduces under realistic clutter, occlusion, and multi-frame dynamics.

\noindent\textbf{Abstract 2D panels:} Following the developmental cascade in human cognitive development~\cite{spelke1990principles,carey2009origin,wellman2001meta}, the regime spans three stages, each requiring the previous as computational input (full diagram in \S\ref{app:templates}). Stage~1 (Object Tracking, 6 templates, 300 probes) tests spatiotemporal individuation: occlusion continuity, trajectory-violation detection, count maintenance, permanence, color-change~\cite{spelke1990principles}. Stage~2 (Goal Attribution, 4 templates, 200 probes) tests inference of goals from motion, corresponding to 6--12 month infant capacities~\cite{woodward1998infants,gergely1995taking}: target inference, detour reasoning, non-goal discrimination, efficient action prediction. Stage~3 (False Belief, 10 templates, 500 probes) tests the canonical ToM capacity~\cite{wimmer1983beliefs,baron1985does,wellman2001meta}: Sally-Anne search and belief reports, reality-belief dissociation, knowledge access, unexpected contents, two-room transfers, perspective divergence, and second-order belief. Each template gets 50 seeded randomized variants for 1{,}000 probes total (generation and seed formula in \S\ref{app:stimuli}). We follow construction conventions of prior work and do not claim the panel set as a contribution; representative probes in Figure~\ref{fig:abstract_examples}; full template descriptions in \S\ref{app:templates}.

\noindent\textbf{Naturalistic 3D game scenes:} \textbf{MuMA-ToM}~\cite{shi2025muma} provides 225 sampled household-activity videos with multiple human-rendered agents performing belief-relevant actions, yielding 900 multiple-choice questions across three types: \textsc{belief} (202), \textsc{belief\_of\_goal} (496), and \textsc{social\_goal} (202). The types are imbalanced; per-model overall accuracy is question-weighted across the 900 items, so it differs from the unweighted mean of per-type aggregates. \textbf{MindPower}~\cite{zhang2025mindpower} provides 492 photorealistic embodied scenarios (36 ThreeDWorld, 456 VirtualHome) with gold annotations across six layers: perception, belief, desire, intention, decision, action. Each layer is scored as the average of BERTScore-F1 and sentence-transformer cosine similarity against the gold reference. MuMA-ToM inputs are eight separate frames per video; MindPower inputs are one storyboard image of up to 12 frames per scenario. Both datasets increase perceptual difficulty along every realistic axis (clutter, occlusion, multi-agent dynamics, multi-frame integration, lighting), inverting the abstract regime. Representative frames in Figure~\ref{fig:naturalistic_examples}.

\begin{table}[t]
  \begin{minipage}[b]{0.485\textwidth}
    \sbox0{%
  \scriptsize
  \setlength{\tabcolsep}{3pt}
  \renewcommand{\arraystretch}{1.05}
  \begin{tabular}{@{}lccccc@{}}
  \toprule
  \textbf{Stage / Type / Layer} & \chipA\,A & \chipBa\,$B_a$ & \chipBp\,$B_p$ & \chipBf\,$B_f$ & \chipC\,C \\
  \midrule
  \multicolumn{6}{@{}l}{\textit{(a) Abstract 2D panels: developmental stages}} \\
  \midrule
  1: Object Tracking & 72.8 & 71.5 & \textbf{96.4} & 93.4 & 89.4 \\
  2: Goal Attribution & 90.9 & 89.4 & \textbf{99.2} & 98.9 & 93.1 \\
  3: False Belief & 73.9 & 75.7 & 88.8 & \textbf{98.4} & 94.5 \\
  \midrule
  \multicolumn{6}{@{}l}{\textit{(b) Naturalistic MuMA-ToM: by question type}} \\
  \midrule
  \textsc{belief} & 58.8 & 64.9 & -- & \textbf{74.7} & 67.7 \\
  \textsc{belief\_of\_goal} & 41.6 & 44.4 & -- & \textbf{52.9} & 50.6 \\
  \textsc{social\_goal} & 37.6 & 37.0 & -- & \textbf{53.6} & 48.8 \\
  \midrule
  \multicolumn{6}{@{}l}{\textit{(c) Naturalistic MindPower: by reasoning layer}} \\
  \midrule
  \textsc{perception} & 19.3 & 21.6 & 26.8 & 24.8 & \textbf{47.4} \\
  \textsc{belief} & 23.4 & 22.7 & 28.8 & 35.5 & \textbf{35.9} \\
  \textsc{desire} & 21.4 & 20.0 & 24.9 & \textbf{29.0} & 28.7 \\
  \textsc{intention} & 21.4 & 19.2 & 25.6 & 28.7 & \textbf{31.3} \\
  \textsc{decision} & 23.3 & 21.2 & 28.1 & 31.3 & \textbf{36.2} \\
  \textsc{action} & 6.1 & 4.9 & 10.2 & 14.4 & \textbf{18.7} \\
  \midrule
  \textit{Overall} & 19.1 & 18.3 & 24.1 & 27.3 & \textbf{33.0} \\
  \bottomrule
  \end{tabular}}%
\begin{minipage}{\linewidth}\centering
  \captionof{table}{\textbf{Aggregate accuracy (\%) by regime and condition}, 7-model means. (b) Pooled over retained responses per type; overall in Table~\ref{tab:permodel_muma}. (c) Weighted over $36$ TDW + $456$ VirtualHome scenarios; $B_a$ averages two parsers (Table~\ref{tab:permodel_mindpower}). \textbf{Bold}: best per row.}
  \label{tab:main}
  \usebox{0}
\end{minipage}
  \end{minipage}\hfill
  \begin{minipage}[b]{0.485\textwidth}
    \sbox0{%
  \scriptsize
  \setlength{\tabcolsep}{3pt}
  \renewcommand{\arraystretch}{1.05}
  \begin{tabular}{@{}lccc@{}}
  \toprule
  \textbf{Tier} & \chipBa\,$\mathrm{PCR}_{a}\,\uparrow$ & \chipBp\,$\mathrm{PCR}_{p}\,\uparrow$ & \chipBf\,$\mathrm{PCR}_{f}\,\uparrow$ \\
  \midrule
  \multicolumn{4}{@{}l}{\textit{(a) Abstract 2D panels, Stage~3}} \\
  \midrule
  Frontier & 0.02 [-.12,\,.15] & 0.96 [.89,\,1.04] & \textbf{1.22} [1.14,\,1.31] \\
  Mid-scale & -1.10 [-1.45,\,-.81] & 0.27 [.12,\,.40] & \textbf{1.01} [.99,\,1.04] \\
  Small-scale & 0.82 [.73,\,.92] & 0.77 [.67,\,.87] & \textbf{1.26} [1.18,\,1.34] \\
  \midrule
  \multicolumn{4}{@{}l}{\textit{(b) Naturalistic MuMA-ToM}} \\
  \midrule
  Frontier & 0.25 [.10,\,.41] & -- & \textbf{1.81} [1.61,\,2.04] \\
  Mid-scale & 0.04 [-.18,\,.27] & -- & \textbf{1.57} [1.32,\,1.86] \\
  Small-scale & 0.53 [.38,\,.69] & -- & \textbf{0.66} [.50,\,.83] \\
  \midrule
  \multicolumn{4}{@{}l}{\textit{(c) Naturalistic MindPower}} \\
  \midrule
  Frontier & -0.06 [-.08,\,-.05] & 0.39 [.37,\,.42] & \textbf{0.49} [.45,\,.53] \\
  Mid-scale & -0.12 [-.15,\,-.09] & 0.42 [.38,\,.46] & \textbf{0.86} [.80,\,.93] \\
  Small-scale & -0.02 [-.04,\,.00] & 0.23 [.20,\,.26] & \textbf{0.51} [.47,\,.56] \\
  \bottomrule
  \end{tabular}}%
\begin{minipage}{\linewidth}\centering
  \captionof{table}{\textbf{PCR by capability tier} (95\% bootstrap CIs). MuMA-ToM: question-weighted. MindPower: scenario-weighted layer-mean over both environments. \textbf{Bold}: best per block.}
  \label{tab:tiers}
  \usebox{0}
\end{minipage}
  \end{minipage}
  \vspace{-0.2cm}
\end{table}

\subsection{The Perceptual Contribution Ratio}
\label{sec:pcr}

The five conditions yield five accuracy numbers per model and benchmark. We collapse them to a single ratio: of the visual-to-text deficit ${\mathrm{Acc}(C) - \mathrm{Acc}(A)}$, what fraction does a scaffold of given fidelity recover?

A VLM processes a stimulus $I$ through perceptual extraction $\pi$, producing internal propositions, followed by reasoning $\rho$ mapping propositions and a question $q$ to an answer: $f_\mathrm{vis}(I, q) = \rho(\pi(I), q)$. In text-only mode, $f_\mathrm{txt}(T, q) = \rho(\tau(T), q)$, where $\tau$ encodes a narration $T$ into the same propositional space. A scaffold $s_x$ at fidelity $x$ augments visual input via token-level fusion: $f_\mathrm{scaffold}(I, s_x, q) = \rho(\pi(I) \oplus \tau(s_x), q)$. Treating $\rho$ as shared across modalities is a simplification; the protocol tests its functional consequence rather than its mechanistic identity.

\begin{definition}[Perceptual Contribution Ratio]
\label{def:pcr}
\begin{equation}
\label{eq:pcr}
\mathrm{PCR}_x = \frac{\mathrm{Acc}(B_x) - \mathrm{Acc}(A)}{\mathrm{Acc}(C) - \mathrm{Acc}(A)},
\end{equation}
where $A$ is visual-only, $B_x$ is image-with-scaffold at fidelity $x \in \{a, p, f\}$, and $C$ is text-only.
\end{definition}

The triplet $(\mathrm{PCR}_a, \mathrm{PCR}_p, \mathrm{PCR}_f)$ traces a dose-response curve. $\mathrm{PCR}_x = 1$ indicates the scaffold closes the gap entirely (purely perceptual deficit); $\mathrm{PCR}_x = 0$ indicates no benefit (purely conceptual deficit); intermediate values give the fraction of the deficit perceptual repair recovers; negative values flag misleading scaffolds; values above $1$ indicate the scaffold-plus-image combination matches or exceeds the text-only ceiling (\S\ref{sec:results}). The metric is undefined when $\mathrm{Acc}(C) = \mathrm{Acc}(A)$; the gaps observed throughout this paper make that case moot.

This functional form is not arbitrary. PCR is the unique metric on triples $(\mathrm{Acc}(A), \mathrm{Acc}(B_x), \mathrm{Acc}(C))$ with $\mathrm{Acc}(C) > \mathrm{Acc}(A)$ that satisfies endpoint calibration, invariance to positive affine reparameterizations, and affine dependence on the scaffolded accuracy (proof in \S\ref{app:pcr_axioms}). Any alternative form must abandon at least one assumption.

\subsection{Models and evaluation}
\label{sec:models}

\noindent\textbf{Model lineup:} We evaluate seven VLMs across three tiers reflecting parameter count and provider positioning. \emph{Frontier} (top-line commercial, $>$100B): GPT-4o (Nov.\ 2024)~\cite{openai2024gpt4o}, Gemini 2.5 Pro Preview~\cite{google2025gemini}, Claude Sonnet 4~\cite{anthropic2025claude}. \emph{Mid-scale} (17B--72B): Qwen2.5-VL-72B-Instruct~\cite{bai2025qwen25vl}, Llama 4 Maverick~\cite{meta2025llama4}. \emph{Small-scale} (sub-10B or mini): GPT-4.1-mini~\cite{openai2025gpt41mini}, Ministral~3 8B~\cite{mistral2025ministral3}. All seven are evaluated on both regimes. Models are accessed through the OpenRouter API with deterministic decoding (temperature 0, max 200 output tokens, single-turn); request parameters and retry logic in \S\ref{app:api}, exact model identifiers in \S\ref{app:addlimitations}, and verbatim prompts for every condition in \S\ref{app:dataset_overview}.

\noindent\textbf{Answer grading:} Responses are normalized (lowercased, whitespace-collapsed) before deterministic grading with no partial credit. For yes/no, the response must begin with the expected token; for numerical, the expected digit must appear in the first four characters; for directional, the response must contain any expected alias, including container color names, where unambiguous; for the reality-versus-belief template, the response is split at ``belief'' with the reality answer before and belief answer after. Identical across all conditions and models; alias lists and split-answer algorithms in \S\ref{app:grading}. MuMA-ToM uses the multiple-choice grading from~\cite{shi2025muma}; MindPower uses the per-layer scoring from~\cite{zhang2025mindpower}.

\noindent\textbf{Statistical methodology:} Confidence intervals are computed by probe-level bootstrap with 10{,}000 iterations, resampling probe indices with replacement and reporting the 2.5th and 97.5th percentiles. PCR CIs use the same bootstrap, recomputing the ratio per iteration. Pairwise condition comparisons use McNemar's test on matched (model, probe) pairs with Bonferroni correction across the 12 protocol comparisons ($\alpha_\mathrm{Bonf} = 0.0042$); for abstract Stage~3, this yields $n = 7 \times 500 = 3{,}500$ paired observations per comparison. The ladder and corruption controls (\S\ref{sec:results}) use the same test and threshold against the same A. Full detail in \S\ref{app:statistics}.

\noindent\textbf{Scale:} The protocol totals roughly $159{,}000$ API calls: $\sim$$111{,}000$ on the abstract regime (including the content ladder, corruption controls, and ablations on $B_\mathrm{vlm}$, $B_\mathrm{vis-stripped}$, chain-of-thought, and self-scaffold reported in \S\ref{app:rebuttal}) and $\sim$$48{,}000$ on the naturalistic regime. Parser-failure responses are excluded from totals.
\begin{table}[t]
  \begin{minipage}[b]{0.485\textwidth}
    \sbox0{%
  \scriptsize
  \setlength{\tabcolsep}{3pt}
  \renewcommand{\arraystretch}{1.05}
  \begin{tabular}{@{}lccccc@{}}
  \toprule
  \textbf{Model} & \chipA\,A & \chipBa\,$B_a$ & \chipBp\,$B_p$ & \chipBf\,$B_f$ & \chipC\,C \\
  \midrule
  GPT-4o~\cite{openai2024gpt4o} & 76.4 & 86.6 & 90.2 & \textbf{99.2} & 91.6 \\
  Gemini 2.5 Pro~\cite{google2025gemini} & 71.8 & 68.4 & 90.4 & 92.6 & \textbf{95.4} \\
  Claude Sonnet 4~\cite{anthropic2025claude} & 78.8 & 73.2 & 97.6 & \textbf{100.0} & 93.2 \\
  \midrule
  Qwen2.5-VL-72B~\cite{bai2025qwen25vl} & 79.2 & 57.0 & 90.0 & \textbf{100.0} & 99.4 \\
  Llama 4 Maverick~\cite{meta2025llama4} & 87.0 & 72.8 & 85.0 & 99.8 & \textbf{100.0} \\
  \midrule
  GPT-4.1-mini~\cite{openai2025gpt41mini} & 70.2 & 80.8 & 80.2 & \textbf{100.0} & 94.8 \\
  Ministral-8B~\cite{mistral2025ministral3} & 53.6 & 91.2 & 88.4 & \textbf{97.4} & 87.0 \\
  \midrule
  \textbf{Mean} & 73.9 & 75.7 & 88.8 & \textbf{98.4} & 94.5 \\
  \bottomrule
  \end{tabular}}%
\begin{minipage}{\linewidth}\centering
  \captionof{table}{\textbf{Per-model accuracy (\%) on abstract 2D Stage~3 (False Belief).} $n{=}500$ probes per model. Models are separated by tier (see \S\ref{sec:models}). \textbf{Bold}: best per row.}
  \label{tab:permodel_abstract}
  \usebox{0}
\end{minipage}
  \end{minipage}\hfill
  \begin{minipage}[b]{0.485\textwidth}
    \sbox0{%
  \scriptsize
  \setlength{\tabcolsep}{3pt}
  \renewcommand{\arraystretch}{1.05}
  \begin{tabular}{@{}lcccc@{}}
  \toprule
  \textbf{Model} & \chipA\,A & \chipBa\,$B_a$ & \chipBf\,$B_f$ & \chipC\,C \\
  \midrule
  GPT-4o~\cite{openai2024gpt4o} & 45.8 & 47.6 & 52.4 & \textbf{56.4} \\
  Gemini 2.5 Pro~\cite{google2025gemini} & 62.3 & 59.1 & \textbf{84.2} & 65.1 \\
  Claude Sonnet 4~\cite{anthropic2025claude} & 33.8 & 43.3 & \textbf{64.9} & 52.8 \\
  \midrule
  Qwen2.5-VL-72B~\cite{bai2025qwen25vl} & 48.1 & 45.1 & \textbf{58.1} & 55.8 \\
  Llama 4 Maverick~\cite{meta2025llama4} & 44.2 & 47.8 & \textbf{55.7} & 50.2 \\
  \midrule
  GPT-4.1-mini~\cite{openai2025gpt41mini} & 41.1 & 46.9 & 51.4 & \textbf{55.8} \\
  Ministral-8B~\cite{mistral2025ministral3} & 37.4 & 41.9 & 39.9 & \textbf{42.1} \\
  \midrule
  \textbf{Mean} & 44.7 & 47.4 & \textbf{58.1} & 54.0 \\
  \bottomrule
  \end{tabular}}%
\begin{minipage}{\linewidth}\centering
  \captionof{table}{\textbf{Per-model accuracy (\%) on MuMA-ToM} ($n{=}900$ probes/model; $B_p$ undefined). Unparsed responses excluded; Mean averages the per-model accuracies. \textbf{Bold}: best per row.}
  \label{tab:permodel_muma}
  \usebox{0}
\end{minipage}
  \end{minipage}
  \vspace{-0.3cm}
\end{table}

\section{Results}
\label{sec:results}

The protocol described in \S\ref{sec:protocol} was applied across the seven-model lineup on both stimulus regimes, totaling roughly $159{,}000$ API calls. We organize the findings around five claims that the dose-response signature licenses, followed by a robustness check. Headline numbers appear in Table~\ref{tab:main}, per-tier PCR in Table~\ref{tab:tiers}, and per-model breakdowns in Tables~\ref{tab:permodel_abstract}, \ref{tab:permodel_muma}, and~\ref{tab:permodel_mindpower}.

\noindent\textbf{Ground-truth scaffolds substantially close the visual-to-text gap on every regime:} On abstract 2D Stage~3, aggregate visual-only accuracy is $73.9\%$ and text-only is $94.5\%$, a gap of $20.6$ pp. The full ground-truth scaffold $B_f$ closes this gap entirely, reaching $98.4\%$, with $\mathrm{PCR}_f = 1.19$ (95\% CI [1.09, 1.30]). The $B_f$-plus-image condition matches or exceeds the text-only ceiling for every tier (Table~\ref{tab:tiers}a): frontier $1.22$, mid-scale $1.01$, small-scale $1.26$. Naturalistic MuMA-ToM shows the same pattern at a larger magnitude: aggregate $\mathrm{PCR}_f = 1.44$, frontier $1.81$, mid-scale $1.57$, small-scale $0.66$ (Table~\ref{tab:tiers}b). These scaffolds match none of the visibility patterns of \S\ref{app:visstrip} (\S\ref{app:muma_audit}), so content without explicit visibility statements closes the gap. Because $B_f$ and C also differ in their text on abstract panels, we do not interpret $\mathrm{PCR} > 1$ further. On MindPower, the picture changes quantitatively but not qualitatively: $B_f$ recovers $59\%$ of a $13.9$ pp visual-to-text gap, with mid-scale strongest at $\mathrm{PCR}_f = 0.86$. On this regime, $B_f$ is an oracle full-information scaffold (\S\ref{sec:conditions}); the leakage-controlled $B_p$ recovers $39\%$ of the gap on the frontier and $42\%$ on mid-scale (Table~\ref{tab:tiers}c). Across all three regimes, every $\mathrm{PCR}_f$ is positive and statistically distinct from zero (Table~\ref{tab:tiers} CIs; McNemar $p_\mathrm{Bonf} < 10^{-4}$ on the binary-scored regimes).

\begin{table}[t]
\centering
\caption{\textbf{Per-model layer-mean accuracy (\%) on MindPower.} $B_a^G$ and $B_a^C$: GPT-4o and Claude Sonnet~4 as automatic parsers. $B_f$: oracle scaffold with the human reference chain across all six layers (\S\ref{sec:conditions}). Tiers: F/M/S = Frontier/Mid-scale/Small-scale. \textbf{Bold}: best per row within each environment.}
\label{tab:permodel_mindpower}
\scriptsize
\setlength{\tabcolsep}{2.4pt}
\renewcommand{\arraystretch}{1.05}
\begin{tabular}{@{}ll cccccc cccccc@{}}
\toprule
& & \multicolumn{6}{c}{\textbf{ThreeDWorld (36 scenarios)}} & \multicolumn{6}{c}{\textbf{VirtualHome (456 scenarios)}} \\
\cmidrule(lr){3-8}\cmidrule(l){9-14}
\textbf{Model} & \textbf{Tier} & \chipA\,A & \chipBa\,$B_a^G$ & \chipBa\,$B_a^C$ & \chipBp\,$B_p$ & \chipBf\,$B_f$ & \chipC\,C & \chipA\,A & \chipBa\,$B_a^G$ & \chipBa\,$B_a^C$ & \chipBp\,$B_p$ & \chipBf\,$B_f$ & \chipC\,C \\
\midrule
GPT-4o~\cite{openai2024gpt4o} & F & 31.6 & 30.7 & 26.9 & 45.1 & \textbf{54.9} & \textbf{54.9} & 19.3 & 17.1 & 16.9 & 23.5 & \textbf{31.7} & 29.5 \\
Gemini 2.5 Pro~\cite{google2025gemini} & F & 35.7 & 34.6 & 33.5 & 41.9 & 46.4 & \textbf{57.1} & 19.4 & 18.4 & 18.3 & 24.5 & 22.8 & \textbf{34.2} \\
Claude Sonnet 4~\cite{anthropic2025claude} & F & 25.5 & 31.3 & 26.3 & 45.4 & 41.6 & \textbf{55.8} & 18.3 & 18.5 & 18.9 & 25.8 & 23.0 & \textbf{37.5} \\
\midrule
Qwen2.5-VL-72B~\cite{bai2025qwen25vl} & M & 28.8 & 29.2 & 26.1 & 49.0 & 44.5 & \textbf{49.1} & 19.4 & 18.2 & 17.3 & 23.4 & 26.8 & \textbf{31.1} \\
Llama 4 Maverick~\cite{meta2025llama4} & M & 24.7 & 28.8 & 25.3 & 44.5 & \textbf{51.9} & 51.2 & 18.4 & 17.2 & 16.7 & 21.7 & \textbf{29.3} & 28.1 \\
\midrule
GPT-4.1-mini~\cite{openai2025gpt41mini} & S & 32.6 & 31.5 & 29.9 & 45.2 & 45.1 & \textbf{56.6} & 19.8 & 18.8 & 18.8 & 23.4 & 25.5 & \textbf{35.3} \\
Ministral-8B~\cite{mistral2025ministral3} & S & 20.2 & 23.3 & 20.4 & 27.7 & 35.9 & \textbf{38.2} & 14.4 & 14.6 & 14.9 & 15.9 & 21.6 & \textbf{25.2} \\
\midrule
\textbf{Mean} & & 28.4 & 29.9 & 26.9 & 42.7 & 45.8 & \textbf{51.9} & 18.4 & 17.5 & 17.4 & 22.6 & 25.8 & \textbf{31.5} \\
\bottomrule
\end{tabular}
\vspace{-0.3cm}
\end{table}

\noindent\textbf{Automatic parsers fail across regimes:} Where ground-truth scaffolds rescue performance, automatic-parser scaffolds do not (Table~\ref{tab:tiers}, $\mathrm{PCR}_a$ column). On abstract 2D Stage~3, the aggregate $\mathrm{PCR}_a$ is $0.09$, with the mid-scale tier at $-1.10$: the parser actively misleads. The small-scale tier is the exception ($\mathrm{PCR}_a = 0.82$). The hand-engineered abstract parser correctly recovers entity identities and positions on $99.4\%$ of probes (\S\ref{app:pertemplate}), so the failure is not at low-level extraction. The deficit is that parser output, even when correct on entity-level facts, lacks the visibility and presence-absence structure false-belief requires. On MindPower we ran two independent parsers: GPT-4o-as-parser yields aggregate $\mathrm{PCR}_a = -0.06$ across tiers and Claude-as-parser yields $-0.08$ (scenario-weighted average of environment PCRs; Table~\ref{tab:permodel_mindpower}). Failure consistency across two frontier parsers, in addition to the hand-engineered parser on abstract panels, is the diagnostic: the bottleneck is not bridged by current visual encoders in zero-shot scene-describer roles.

\noindent\textbf{The dose-response is monotonic in scaffold fidelity:} Where all three scaffold conditions are defined (abstract 2D Stage~3 and MindPower), accuracy increases monotonically from $A$ through $B_a$ through $B_p$ through $B_f$, the only exception being $B_a$, which falls slightly below $A$ on MindPower for the reasons above. Abstract Stage~3: $A=73.9 \to B_a=75.7 \to B_p=88.8 \to B_f=98.4$. MindPower (overall layer-mean): $A=19.1 \to B_a=18.3 \to B_p=24.1 \to B_f=27.3$. The ordering rules out non-perceptual explanations: if scaffolds helped through token-count or format-priming, $B_a$ would close more of the gap and $B_p$ would not be reliably below $B_f$. Excluding the automatic parser, the ordering $A < B_p < B_f$ holds per model for $6$ of $7$ models on abstract Stage~3 (Table~\ref{tab:permodel_abstract}); the exception is Llama~4 Maverick, whose $B_p$ ($85.0\%$) falls below its high visual-only baseline ($87.0\%$). A stripped variant $B_\mathrm{vis-stripped}$ removes the explicit visibility and absence statements that match nine regex patterns from $B_f$ (\S\ref{app:visstrip}; $8.9\%$ of characters). On the six models evaluated, accuracy is $90.0\%$ and $\mathrm{PCR}_\mathrm{vis-stripped} = 0.69$, recovering most of the gap. The one outlier is Gemini 2.5 Pro ($54.4\%$, against $88.0\%$ or above for every other model), indicating model-specific reliance on explicit absence cues (Table~\ref{tab:vis_stripped_supp}).

\begin{table}[t]
  \begin{minipage}[t]{0.485\textwidth}
    \sbox0{%
  \scriptsize
  \setlength{\tabcolsep}{3pt}
  \renewcommand{\arraystretch}{1.05}
  \begin{tabular}{@{}llcc@{}}
  \toprule
  \textbf{Condition} & \textbf{Content added} & \textbf{Acc} & $\Delta$ \vs A \\
  \midrule
  \chipA\,A & none & 73.9 & -- \\
  $B_\mathrm{identity}$ & per-panel entity lists & 74.0 & \up{+0.1} \\
  $B_\mathrm{static}$ & + per-panel positions & 76.7 & \up{+2.8} \\
  \chipBp\,$B_p$ & + cross-frame events, linkage & 88.8 & \up{+14.9} \\
  $B_\mathrm{vis\text{-}stripped}$ & $B_f$ minus visibility statements & 90.0$^\dagger$ & \up{+12.8}$^\dagger$ \\
  \chipBf\,$B_f$ & + visibility and absence & \textbf{98.4} & \up{+24.5} \\
  \chipC\,C & text narration, no image & 94.5 & \up{+20.6} \\
  \midrule
  \multicolumn{4}{@{}l}{\textit{Length-matched corruptions of $B_p$}} \\
  \midrule
  $D_\mathrm{mismatch}$ & another variant's scaffold & 74.7 & \up{+0.8} \\
  $D_\mathrm{noise}$ & 30\% of names replaced & 79.5 & \up{+5.6} \\
  $D_\mathrm{shuffled}$ & panel sentences reordered & 80.6 & \up{+6.7} \\
  \bottomrule
  \end{tabular}}%
\begin{minipage}{\linewidth}\centering
  \captionof{table}{\textbf{Scaffold-content ladder and corruption controls} on Stage~3 (7 models $\times$ 500 probes). No rung below $B_f$ states a belief. $^\dagger$Six models; $\Delta$ \vs their A (77.2).}
  \label{tab:ladder}
  \usebox{0}
\end{minipage}

  \end{minipage}\hfill
  \begin{minipage}[t]{0.485\textwidth}
    \sbox0{%
  \scriptsize
  \setlength{\tabcolsep}{3pt}
  \renewcommand{\arraystretch}{1.05}
  \begin{tabular}{@{}lccc@{}}
  \toprule
  \textbf{Variant} & \chipA\,A & \chipC\,C & \chipBf\,$B_f$ \\
  \midrule
  \multicolumn{4}{@{}l}{\textit{(a) Prompt paraphrase}} \\
  \midrule
  Original & 73.9 & 95.7 & 98.6 \\
  Restructured & 70.1\,\dn{$-$3.8} & 95.2\,\dn{$-$0.5} & 93.9\,\dn{$-$4.7} \\
  Vocabulary & 71.9\,\dn{$-$2.0} & 94.8\,\dn{$-$0.9} & 97.1\,\dn{$-$1.5} \\
  Distractor / adversarial & 69.9\,\dn{$-$4.0} & 97.2\,\up{+1.5} & 97.6\,\dn{$-$1.0} \\
  \midrule
  \multicolumn{4}{@{}l}{\textit{(b) Prompting strategy (visual only)}} \\
  \midrule
  Chain-of-thought & 76.6\,\dn{$-$1.7} & -- & -- \\
  Self-scaffold & 58.5\,\dn{$-$18.7} & -- & -- \\
  \bottomrule
  \end{tabular}}%
\begin{minipage}{\linewidth}\centering
  \captionof{table}{\textbf{Prompt sensitivity} on Stage~3. A: 7 models (paraphrases on 200 probes each); C, $B_f$: 6 models. Last row: distractor (A), adversarial (C, $B_f$). (b) $\Delta$ \vs A on the same models.}
  \label{tab:prompt}
  \usebox{0}
\end{minipage}

  \end{minipage}
  \vspace{-0.3cm}
\end{table}

\newcommand{\qok}{\textcolor{deltaup}{\ding{51}}}
\newcommand{\qno}{\textcolor{deltadn}{\ding{55}}}
\begin{figure}[t]
  \centering
  \scriptsize
  \begin{minipage}[c]{0.52\linewidth}
    \includegraphics[width=\linewidth]{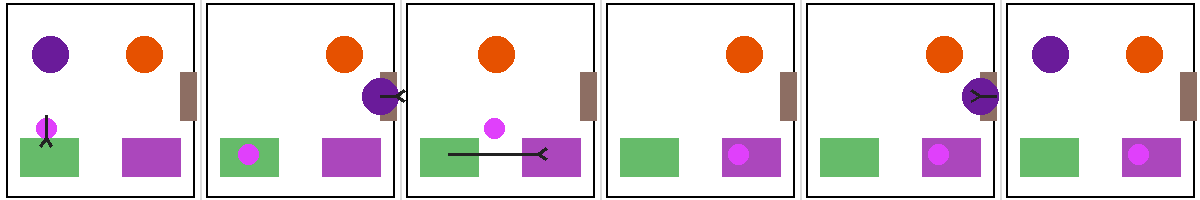}
  \end{minipage}\hfill
  \begin{minipage}[c]{0.455\linewidth}
    \raggedright
    \textbf{(a)} \textsc{sally\_anne\_reality}, GPT-4o. \emph{Where is the pink object actually located in the final panel?} (correct: right)\\[3pt]
    \legendchip{condA}{A}\,``left''\,\qno\quad
    \legendchip{condBperc}{$B_p$}\,``right''\,\qok\quad
    \legendchip{condBfull}{$B_f$}\,``right''\,\qok\quad
    \legendchip{condC}{C}\,``right''\,\qok\\[3pt]
    Template accuracy: A 74.9, $B_p$ 98.9, $B_f$ 99.4, C 100.0.
  \end{minipage}\\[5pt]
  \begin{minipage}[c]{0.38\linewidth}
    \includegraphics[width=\linewidth]{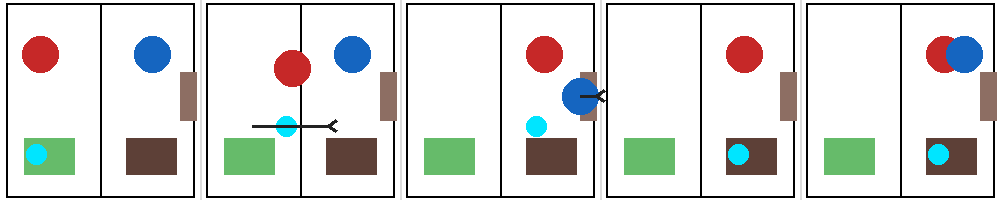}
  \end{minipage}\hfill
  \begin{minipage}[c]{0.60\linewidth}
    \raggedright
    \textbf{(b)} \textsc{two\_rooms}, GPT-4o. \legendchip{condC}{C} \emph{red starts in the left room with the cyan object. blue is in the right room. red carries the object into the right room while blue is watching. blue then leaves. While blue is gone, red places the object in the right brown container. blue returns later. Where does the blue character think the cyan object is, left or right?} (correct: right)\quad Output: ``left''\,\qno\\[3pt]
    Template accuracy: A 56.0, $B_p$ 65.7, $B_f$ 98.9, C 78.6.
  \end{minipage}
  \vspace{-0.1cm}
  \caption{\textbf{Qualitative examples with verbatim outputs.} \textbf{(a)}~Visual-only failure recovered by scaffolds: on a reality question, the model reports the pre-transfer location. \textbf{(b)}~Failure at the text-only ceiling: the returning agent saw the object carried into the right room, yet the model applies ``absent agent, therefore old location''. Template accuracy over 350 trials; prompts in \S\ref{app:dataset_overview}.}
  \label{fig:qualitative}
  \vspace{-0.3cm}
\end{figure}

\noindent\textbf{The recovery tracks structured event content, not text volume:} Two rungs below $B_p$ separate perceptual repair from offloaded state tracking (Table~\ref{tab:ladder}; \S\ref{app:ladder}): $B_\mathrm{identity}$ lists the entities in each panel without positions, and $B_\mathrm{static}$ adds per-panel positions with no events, linkage, or visibility statements. On aggregate these add little ($+0.1$ pp, $p = .84$; $+2.8$ pp, $p = .0033$), the largest increment ($+12.1$ pp to $B_p$) comes from cross-frame event structure, which carries no mental-state content, and $B_f$, which adds observable access facts, contributes the remaining $9.6$ pp. Per model the picture is heterogeneous: static positions alone recover most or all of the $B_p$ gain for GPT-4o, GPT-4.1-mini, and Ministral-8B, but hurt Claude Sonnet~4, Llama~4 Maverick, and Qwen2.5-VL-72B, whose $B_\mathrm{static}$ failures remain reality bias in $47$ of $50$ inspected cases. Three length-matched corruptions of $B_p$ (Table~\ref{tab:ladder}; \S\ref{app:corrupt}) recover less than half of the $14.9$ pp that $B_p$ adds on aggregate ($p \approx 4 \times 10^{-63}$ for $B_p$). $D_\mathrm{mismatch}$, the strongest corruption, gives no significant gain ($p = .38$); the two weaker ones give small but significant gains ($+5.6$ and $+6.7$ pp, both $p < 10^{-8}$), much of it from Ministral-8B. Per model, Gemini 2.5 Pro falls below A under every corruption and Claude Sonnet~4 retains little of the benefit, whereas GPT-4o retains nearly all of it. Nor does text volume: the VLM-parser scaffolds are the longest (mean $343$ tokens) yet do not improve on A (\S\ref{app:length}); whether the benefit requires image-bound content remains open (\S\ref{sec:limitations}).

\noindent\textbf{The dominant visual-only error mode is reality bias:} We categorized all $915$ visual-only errors on abstract Stage~3 ($7$ models $\times$ $500$ probes) using a deterministic rule (\S\ref{app:errorcats}). An error is a \emph{reality-bias} error if the model reports the object's veridical current location rather than the location consistent with the absent agent's belief. Reality-bias errors account for $549$ of $915$ visual-only errors ($60.0\%$); the remaining $40\%$ are uncategorized. The reality-bias share is invariant across tiers within $5$ pp. The bias is not a failure of belief reasoning in the abstract: under the same probes, text-only accuracy is $94.5\%$. It appears only when the question is conveyed visually and disappears when the visibility-and-absence content is provided in the scaffold. The mechanism is a perceptual front-end that encodes \emph{what} an object is and \emph{where} it currently is but not \emph{who was present when it moved}, leaving the inferential pathway with insufficient state to dissociate reality from belief. Figure~\ref{fig:qualitative} shows a visual-only failure recovered by scaffolds and a residual failure at the text-only ceiling; text-side errors exist, and our claim concerns the $20.6$ pp gap, not perfect text-side reasoning.

\noindent\textbf{The findings are robust to prompting choices:} Visual-only accuracy stays within $4.0$ pp of the original across three question paraphrases, and all six paraphrases of C and $B_f$, including an adversarial insertion, exceed $93\%$ (Table~\ref{tab:prompt}). Prompting for more inference does not substitute for supplied scene state: chain-of-thought changes visual-only accuracy by $-1.7$ pp on average and self-scaffolding by $-18.7$ pp. Chain-of-thought helps Claude Sonnet~4 ($78.8 \to 90.6$) but not past its own $B_p$ ($97.6$): perception and reasoning are intertwined in implementation, and our claim is functional, as the deficit closes when scene state is supplied, not when inference is prompted (\S\ref{app:cot}).

\section{Discussion}
\label{sec:discussion}

\noindent\textbf{Scene-state extraction is the dominant bottleneck:} The findings converge on a single picture. When the scene is given in pixels, frontier VLMs fail false-belief tasks by margins commonly attributed to a missing belief-reasoning capacity. When the same problem is given in text, the same models succeed. Given pixels plus the structured perceptual content the front-end should have produced, performance reaches or exceeds the text-only ceiling on abstract panels and MuMA-ToM, and recovers part of it on MindPower. Under the evaluated protocol, the deficit lies predominantly not in the inferential pathway that operates on scene representations, which the text-only ceiling shows is largely intact, but in the front-end that converts pixels to scene representations. Reality bias names the failure: the front-end encodes object identity and current location but not who was present at each moment, leaving the inferential pathway with the wrong input.

\noindent\textbf{Reconciliation with prior accounts:} Recent cognitive or representational diagnoses of visual ToM failures~\cite{liu2026seeing,chen2024vlms_see,tong2024eyes,hu2025reevaluating} target single-image or static tasks with spatially localized evidence, so they can coexist with our claim. The protocol distinguishes the two diagnoses per task: a dose-response with $\mathrm{PCR}_p$ or $\mathrm{PCR}_f$ near $1$ localizes the deficit to perceptual extraction, and a flat one would localize it to inference. For sequential multi-panel and multi-frame ToM, the signature consistently localizes to perception, and it supplies the measurement that model-based approaches~\cite{rabinowitz2018tomnet,jin2024mmtom} skip by taking scene structure as given (\S\ref{app:competing}).

\noindent\textbf{Architectural implications:} If the bottleneck is in front-end scene-state extraction rather than belief inference, the first priority is better perceptual representations rather than new reasoning modules. The specific competences current encoders fail to produce are object-level entity tracking with persistent identity across frames, and presence-absence registers indexed to entities at each frame. We predict that VLM architectures with explicit object slots, cross-frame correspondence, and visibility tracking, pretrained for per-entity state extraction rather than global scene captioning, would close most of the visual-to-text gap on existing ToM benchmarks without belief-specific training objectives.

\section{Limitations}
\label{sec:limitations}

We evaluate two stimulus regimes and seven black-box API models; in-the-wild video and open-weight object-centric models are untested, and the claims are functional, not mechanistic. The perception/reasoning boundary follows our operational definition (\S\ref{sec:protocol}): cross-frame event linking is an intermediate representation, and a definition assigning it to reasoning would move that increment accordingly. The corruption controls are model-dependent and partly weak by construction, and $D_\mathrm{mismatch}$ may net useful template cues against misleading scene content, so they argue against text volume without settling whether the benefit requires image-bound content; jointly varying image-text alignment is a natural next test (\S\ref{app:addlimitations}).

\section{Conclusion}
\label{sec:conclusion}

We proposed a five-condition dissociation protocol that holds the inferential task constant while varying perceptual fidelity. Across seven VLMs and roughly $159{,}000$ API calls, the visual ToM deficit is predominantly perceptual: ground-truth scaffolds substantially close the visual-to-text gap on every regime, automatic parsers fail in aggregate, the recovery tracks cross-frame event structure, and the dominant visual-only error is reality bias. Under the evaluated protocol, structured visual scene-state extraction, rather than belief inference, is the dominant bottleneck for visual ToM.

\clearpage

\section*{Acknowledgements}
Mohamed Rayan Barhdadi has been supported by the Student Research Experience Grant (SREG-ECEN-2026) from the Office of Research and Graduate Studies, TAMU-Q. Syed Talal Wasim has been supported by the 2026 Google PhD Fellowship in Machine Perception and, together with Juergen Gall, by the ERC Consolidator Grant FORHUE (101044724). Juergen Gall has additionally been supported by the Federal Ministry of Research, Technology and Space (BMFTR) under grant no.~16IS22094A WEST-AI.

\bibliographystyle{plain}
\bibliography{main}

\begin{thebibliography}{10}

\bibitem{anthropic2025claude}
{Anthropic}.
\newblock {Claude Sonnet 4}, 2025.
\newblock \url{https://www.anthropic.com/claude}.

\bibitem{apperly2009two}
Ian~A Apperly and Stephen~A Butterfill.
\newblock Do humans have two systems to track beliefs and belief-like states?
\newblock {\em Psychological Review}, 116(4):953--970, 2009.

\bibitem{bai2025qwen25vl}
Shuai Bai, Keqin Chen, Xuejing Liu, Jialin Wang, Wenbin Ge, Sibo Song, Kai
  Dang, Peng Wang, Shijie Wang, Jun Tang, Humen Zhong, Yuanzhi Zhu, Mingkun
  Yang, Zhaohai Li, Jianqiang Wan, Pengfei Wang, Wei Ding, Zheren Fu, Yiheng
  Xu, Jiabo Ye, Xi~Zhang, Tianbao Xie, Zesen Cheng, Hang Zhang, Zhibo Yang,
  Haiyang Xu, Junyang Lin, et~al.
\newblock {Qwen2.5-VL} technical report.
\newblock {\em arXiv preprint arXiv:2502.13923}, 2025.

\bibitem{baillargeon2010false}
Ren{\'e}e Baillargeon, Rose~M Scott, and Zijing He.
\newblock False-belief understanding in infants.
\newblock {\em Trends in Cognitive Sciences}, 14(3):110--118, 2010.

\bibitem{baron1985does}
Simon Baron-Cohen, Alan~M Leslie, and Uta Frith.
\newblock Does the autistic child have a ``theory of mind''?
\newblock {\em Cognition}, 21(1):37--46, 1985.

\bibitem{carey2009origin}
Susan Carey.
\newblock {\em The Origin of Concepts}.
\newblock Oxford University Press, 2009.

\bibitem{gandhi2023understanding}
Kanishk Gandhi, Jan-Philipp Fr{\"a}nken, Tobias Gerstenberg, and Noah~D
  Goodman.
\newblock Understanding social reasoning in language models with language
  models.
\newblock In {\em Advances in Neural Information Processing Systems},
  volume~37, 2023.

\bibitem{chen2024vlms_see}
Qingying Gao, Yijiang Li, Haiyun Lyu, Haoran Sun, Dezhi Luo, and Hokin Deng.
\newblock Vision language models see what you want but not what you see.
\newblock {\em arXiv preprint arXiv:2410.00324}, 2024.
\newblock ICLR 2025 Workshop on Bidirectional Human-AI Alignment (BiAlign).

\bibitem{gergely1995taking}
Gy{\"o}rgy Gergely, Zolt{\'a}n N{\'a}dasdy, Gergely Csibra, and Szilvia
  B{\'\i}r{\'o}.
\newblock Taking the intentional stance at 12 months of age.
\newblock {\em Cognition}, 56(2):165--193, 1995.

\bibitem{google2025gemini}
{Google DeepMind}.
\newblock {Gemini 2.5 Pro}, 2025.
\newblock \url{https://deepmind.google/technologies/gemini/}.

\bibitem{heider1944experimental}
Fritz Heider and Marianne Simmel.
\newblock An experimental study of apparent behavior.
\newblock {\em The American Journal of Psychology}, 57(2):243--259, 1944.

\bibitem{hong2025motionbench}
Wenyi Hong, Yean Cheng, Zhuoyi Yang, Weihan Wang, Lefan Wang, Xiaotao Gu, Shiyu
  Huang, Yuxiao Dong, and Jie Tang.
\newblock Motionbench: Benchmarking and improving fine-grained video motion
  understanding for vision language models.
\newblock In {\em Proceedings of the IEEE/CVF Conference on Computer Vision and
  Pattern Recognition (CVPR)}, pages 8450--8460, 2025.

\bibitem{hu2025reevaluating}
Jennifer Hu, Felipe Sosa, and Tomer Ullman.
\newblock Re-evaluating theory of mind evaluation in large language models.
\newblock {\em Philosophical Transactions of the Royal Society B}, 2025.

\bibitem{jin2024mmtom}
Chuanyang Jin, Yutong Wu, Jing Cao, Jiannan Xiang, Yen-Ling Kuo, Zhiting Hu,
  Tomer Ullman, Antonio Torralba, Joshua~B Tenenbaum, and Tianmin Shu.
\newblock {MMToM-QA}: Multimodal theory of mind question answering.
\newblock {\em arXiv preprint arXiv:2401.08743}, 2024.

\bibitem{kim2023fantom}
Hyunwoo Kim, Melanie Sclar, Xuhui Zhou, Ronan Le~Bras, Gunhee Kim, Yejin Choi,
  and Maarten Sap.
\newblock {FANToM}: A benchmark for stress-testing machine theory of mind in
  interactions.
\newblock In {\em Proceedings of the 2023 Conference on Empirical Methods in
  Natural Language Processing (EMNLP)}, pages 14397--14413, 2023.

\bibitem{kosinski2024evaluating}
Michal Kosinski.
\newblock Evaluating large language models in theory of mind tasks.
\newblock {\em Proceedings of the National Academy of Sciences}, 121(45), 2024.

\bibitem{leslie1987pretense}
Alan~M Leslie.
\newblock Pretense and representation: The origins of ``theory of mind''.
\newblock {\em Psychological Review}, 94(4):412--426, 1987.

\bibitem{li2024cogdevelop2k}
Yijiang Li, Qingying Gao, Haoran Sun, Haiyun Lyu, Dezhi Luo, and Hokin Deng.
\newblock Cogdevelop2k: Reversed cognitive development in multimodal large
  language models.
\newblock {\em arXiv preprint arXiv:2410.10855}, 2024.

\bibitem{liu2026seeing}
Zhining Liu, Ziyi Chen, Hui Liu, Chen Luo, Xianfeng Tang, Suhang Wang, Jingying
  Zeng, Zhenwei Dai, Zhan Shi, Tianxin Wei, Hanqing Lu, Benoit Dumoulin, and
  Hanghang Tong.
\newblock Seeing but not believing: Probing the disconnect between visual
  attention and answer correctness in {VLMs}.
\newblock In {\em International Conference on Learning Representations (ICLR)},
  2026.

\bibitem{meta2025llama4}
{Meta AI}.
\newblock {Llama 4}, 2025.
\newblock \url{https://ai.meta.com/llama/}.

\bibitem{mistral2025ministral3}
{Mistral AI}.
\newblock {Ministral 3 8B} (2512), 2025.
\newblock Released December 2, 2025.
  \url{https://openrouter.ai/mistralai/ministral-8b-2512}.

\bibitem{onishi2005do}
Kristine~H Onishi and Ren{\'e}e Baillargeon.
\newblock Do 15-month-old infants understand false beliefs?
\newblock {\em Science}, 308(5719):255--258, 2005.

\bibitem{openai2024gpt4o}
{OpenAI}.
\newblock {GPT-4o}, 2024.
\newblock \url{https://openai.com/gpt-4o}.

\bibitem{openai2025gpt41mini}
{OpenAI}.
\newblock {GPT-4.1 mini}, 2025.
\newblock \url{https://openai.com/index/gpt-4-1/}.

\bibitem{premack1978does}
David Premack and Guy Woodruff.
\newblock Does the chimpanzee have a theory of mind?
\newblock {\em Behavioral and Brain Sciences}, 1(4):515--526, 1978.

\bibitem{rabinowitz2018tomnet}
Neil Rabinowitz, Frank Perbet, Francis Song, Chiyuan Zhang, S.~M.~Ali Eslami,
  and Matthew Botvinick.
\newblock Machine theory of mind.
\newblock In {\em Proceedings of the International Conference on Machine
  Learning (ICML)}, pages 4218--4227, 2018.

\bibitem{sap2022neural}
Maarten Sap, Ronan LeBras, Daniel Fried, and Yejin Choi.
\newblock Neural theory-of-mind? {On} the limits of social intelligence in
  large {LMs}.
\newblock {\em arXiv preprint arXiv:2210.13312}, 2022.

\bibitem{shapira2024clever}
Natalie Shapira, Mosh Levy, Seyed~Hossein Alavi, Xuhui Zhou, Yejin Choi, Yoav
  Goldberg, Maarten Sap, and Vered Shwartz.
\newblock Clever hans or neural theory of mind? stress testing social reasoning
  in large language models.
\newblock {\em arXiv preprint arXiv:2305.14763}, 2024.

\bibitem{shi2025muma}
Haojun Shi, Suyu Ye, Xinyu Fang, Chuanyang Jin, Leyla Isik, Yen-Ling Kuo, and
  Tianmin Shu.
\newblock {MuMA-ToM}: Multi-modal multi-agent theory of mind.
\newblock In {\em Proceedings of the AAAI Conference on Artificial Intelligence
  (AAAI)}, 2025.

\bibitem{shu2021agent}
Tianmin Shu, Abhishek Bhandwaldar, Chuang Gan, Kevin~A Smith, Shari Liu, Dan
  Gutfreund, Elizabeth Spelke, Joshua~B Tenenbaum, and Tomer~D Ullman.
\newblock {AGENT}: A benchmark for core psychological reasoning.
\newblock In {\em Proceedings of the 38th International Conference on Machine
  Learning (ICML)}, pages 9614--9625, 2021.

\bibitem{spelke1990principles}
Elizabeth~S Spelke.
\newblock Principles of object perception.
\newblock {\em Cognitive Science}, 14(1):29--56, 1990.

\bibitem{spelke2000core}
Elizabeth~S Spelke.
\newblock Core knowledge.
\newblock {\em American Psychologist}, 55(11):1233--1243, 2000.

\bibitem{strachan2024testing}
James W.~A. Strachan, Dalila Albergo, Giulia Borghini, Oriana Pansardi, Eugenio
  Scaliti, Saurabh Gupta, Krati Saxena, Alessandro Rufo, Stefano Panzeri, Guido
  Manzi, Michael S.~A. Graziano, and Cristina Becchio.
\newblock Testing theory of mind in large language models and humans.
\newblock {\em Nature Human Behaviour}, 8:1285--1295, 2024.

\bibitem{tong2024eyes}
Shengbang Tong, Zhuang Liu, Yuexiang Zhai, Yi~Ma, Yann LeCun, and Saining Xie.
\newblock Eyes wide shut? exploring the visual shortcomings of multimodal
  {LLMs}.
\newblock In {\em Proceedings of the IEEE/CVF Conference on Computer Vision and
  Pattern Recognition (CVPR)}, pages 9568--9578, 2024.

\bibitem{ullman2023large}
Tomer Ullman.
\newblock Large language models fail on trivial alterations to theory-of-mind
  tasks.
\newblock {\em arXiv preprint arXiv:2302.08399}, 2023.

\bibitem{wellman2001meta}
Henry~M Wellman, David Cross, and Julanne Watson.
\newblock Meta-analysis of theory-of-mind development: The truth about false
  belief.
\newblock {\em Child Development}, 72(3):655--684, 2001.

\bibitem{wimmer1983beliefs}
Heinz Wimmer and Josef Perner.
\newblock Beliefs about beliefs: Representation and constraining function of
  wrong beliefs in young children's understanding of deception.
\newblock {\em Cognition}, 13(1):103--128, 1983.

\bibitem{woodward1998infants}
Amanda~L Woodward.
\newblock Infants selectively encode the goal object of an actor's reach.
\newblock {\em Cognition}, 69(1):1--34, 1998.

\bibitem{zhang2025mindpower}
Ruoxuan Zhang, Qiyun Zheng, Zhiyu Zhou, Ziqi Liao, Siyu Wu, Jian-Yu Jiang-Lin,
  Bin Wen, Hongxia Xie, Jianlong Fu, and Wen-Huang Cheng.
\newblock Mindpower: Enabling theory-of-mind reasoning in {VLM}-based embodied
  agents.
\newblock In {\em Proceedings of the IEEE/CVF Conference on Computer Vision and
  Pattern Recognition (CVPR)}, 2026.

\end{thebibliography}


\clearpage
\appendix

\appendix

\setcounter{section}{0}
\setcounter{figure}{0}
\setcounter{table}{0}
\setcounter{equation}{0}
\renewcommand{\thesection}{\Alph{section}}
\renewcommand{\thefigure}{S\arabic{figure}}
\renewcommand{\thetable}{S\arabic{table}}
\renewcommand{\theequation}{S\arabic{equation}}

\begin{center}
\Large\textbf{Supplementary Material}\\[4pt]
\large\textbf{Perception, Not Reasoning, Limits Visual Theory of Mind}
\end{center}

\condlegend

\smallskip

\noindent This supplementary is organized around a single evidentiary chain: visual false-belief failures persist when models must extract state from pixels, but largely disappear when the missing perceptual state is supplied as text. Section~\ref{app:implementation} fixes the protocol and accounting. Section~\ref{app:extended} shows the full stage and template breakdowns. Section~\ref{app:templates} documents the complete task family. Section~\ref{app:statistics} explains the matched uncertainty estimates and gives the axiomatic characterization of PCR. Section~\ref{app:competing} relates the claim to competing accounts. Section~\ref{app:addlimitations} states the remaining caveats. Section~\ref{app:rebuttal} closes alternative explanations with targeted ablations, including VLM-as-parser, visibility stripping, paraphrases, chain-of-thought, self-scaffold baselines, the scaffold-content ladder, length-matched corruption controls, a scaffold-length analysis, and an audit of the MuMA-ToM scaffolds. Section~\ref{app:dataset_overview} gives verbatim prompts and outputs, including failures at the text-only ceiling.

\section{Implementation details}
\label{app:implementation}

\paragraph{Reproducibility map.} Every protocol element in the main text has a corresponding entry here: stimulus generation and seed formula (\S\ref{app:stimuli}); automatic parser (\S\ref{app:parser}); API request parameters, retry logic, and call accounting (\S\ref{app:api}); answer grading (\S\ref{app:grading}); error categorization (\S\ref{app:errorcats}); prompt paraphrases (\S\ref{app:paraphrases}); statistical tests (\S\ref{app:statistics}); exact model identifiers (\S\ref{app:addlimitations}); ablation scaffolds and controls (\S\ref{app:rebuttal}); and verbatim prompts with model outputs for every condition (\S\ref{app:dataset_overview}). The evaluation harness, stimulus generator, and complete logged responses are released at the URL given in the abstract.

\subsection{Stimulus generation}
\label{app:stimuli}

All 1,000 stimuli are generated programmatically using PIL (Python Imaging Library). Each panel measures $200 \times 200$ pixels. Multi-panel stimuli are created by concatenating panels horizontally, yielding images of width $200 \times N_\mathrm{panels}$ where $N_\mathrm{panels}$ is the panel count (4--6 depending on template). Adjacent panels are separated by a 2-pixel light gray ($\texttt{\#DDDDDD}$) divider.

\paragraph{Entity rendering parameters.}
Characters (agents) are rendered as filled circles with radius 18\,px. Objects are smaller filled circles with radius 10\,px (12\,px for Stage~1 templates). Containers are filled rectangles of $58 \times 38$\,px. Obstacles and doors are gray (\texttt{\#888888}) or brown (\texttt{\#8D6E63}) rectangles. Room boundaries are drawn as 2-pixel black outlines.

\paragraph{Color palettes.}
Six agent colors: blue (\texttt{\#1565C0}), red (\texttt{\#C62828}), green (\texttt{\#2E7D32}), purple (\texttt{\#6A1B9A}), orange (\texttt{\#E65100}), teal (\texttt{\#00838F}). Five object colors: yellow (\texttt{\#FFD600}), amber (\texttt{\#FF6F00}), lime (\texttt{\#76FF03}), pink (\texttt{\#E040FB}), cyan (\texttt{\#00E5FF}). Six container colors: light-blue (\texttt{\#42A5F5}), coral (\texttt{\#EF5350}), brown (\texttt{\#5D4037}), mint (\texttt{\#66BB6A}), violet (\texttt{\#AB47BC}), peach (\texttt{\#FF7043}). For each probe variant, character colors, object colors, and container colors are sampled without replacement from their respective palettes using a seeded random number generator (seed = $10{,}000 \times (\text{template\_index} + 1) + \text{variant\_index}$), ensuring full reproducibility.

\paragraph{Zero text constraint.}
No text labels, panel numbers, character names, or narration appear in any stimulus image.

\subsection{Automatic vision parser}
\label{app:parser}

The parser operates on the saved PNG image with no access to generation metadata.

\paragraph{Foreground masking.}
A pixel is classified as foreground if it is not white ($\text{RGB} > 210$ on all channels), not black ($< 50$ on all channels), not gray (channel spread $< 30$ and mean between 80--210), and not door-brown ($|\text{R} - 141| < 25 \wedge |\text{G} - 110| < 25 \wedge |\text{B} - 99| < 25$).

\paragraph{Connected component extraction.}
Flood-fill BFS from each unvisited foreground pixel, expanding to 4-connected neighbors within $\Delta_\mathrm{RGB} < 60$ from the seed pixel. Components smaller than 40 pixels are discarded. Components are capped at 6,000 pixels to prevent runaway fills.

\paragraph{Entity classification.}
Each blob is classified by area and circularity (area / bounding-box area):
\begin{itemize}[nosep,leftmargin=*]
\item \textbf{Agent}: circular ($0.55 \leq \text{circularity} \leq 0.90$, aspect ratio $0.65$--$1.35$) and area 400--2,000\,px.
\item \textbf{Object}: circular and area 100--550\,px.
\item \textbf{Container}: non-circular and area $\geq 1{,}500$\,px.
\end{itemize}
Color identity is assigned by nearest-neighbor matching against the known palette (Euclidean distance in RGB, threshold 80).

\paragraph{Event inference.}
Agent presence is compared across consecutive panels. If an agent present in panel $k$ is absent in panel $k{+}1$, an exit event is inferred. Object position (left/right of panel center) is tracked to infer transfer events.

\paragraph{Scaffold serialization.}
Parsed structure is serialized as: ``Panel $k$: agents: [name] agent ([side]-[vertical]). objects: [name] object ([side]-[vertical]). containers: [name] container ([side]-[vertical]).''

\subsection{API call protocol}
\label{app:api}

All models are accessed via the OpenRouter API (\texttt{https://openrouter.ai/api/v1/chat/\\completions}).

\paragraph{Request parameters.}
Temperature: 0.0. Max tokens: 200. Single-turn conversation (one user message). Images are included as base64-encoded PNG in the \texttt{image\_url} field for conditions \chipA\,A, \chipBa\,$B_a$, \chipBp\,$B_p$, \chipBf\,$B_f$, and all controls. Condition \chipC\,C sends text only (no image).

\paragraph{Retry logic.}
HTTP 429 (rate limit): exponential backoff starting at 30\,s. HTTP 5xx: retry after 15\,s. Maximum 3 retries per call. Failed calls are recorded as incorrect with an \texttt{ERROR} response prefix.

\paragraph{Parallelization.}
Multiple API keys support parallel workers. Task queue distributes probe-model-condition triples. Results are checkpointed every 100 completions.

\paragraph{Call accounting and cost.}
The paper-facing benchmark used 108,500 API calls: 60,200 for the Abstract 2D panels experiment (five conditions, three adversarial controls, three prompt paraphrases) and 48,300 for the naturalistic 3D game-scene evaluations on MuMA-ToM and MindPower. The corresponding paper-facing cost was \$319.83: \$65.69 for the Abstract 2D panels component and \$254.14 for MuMA-ToM and MindPower. With the 38,500 Abstract 2D panels ablation calls in \S\ref{app:rebuttal} (\$50.00) and 12,000 further calls (\$15.73) for the scaffold-content ladder, the extension of the corruption controls to Ministral-8B, and a re-run of Condition~A to check for API drift (\S\ref{app:ladder}, \S\ref{app:corrupt}; the corruption runs on the other six models are already included in the totals above), the full audit trail comprises ${\sim}159{,}000$ API calls and \$385.56. The Condition~A re-run gave $73.5\%$ (2,572/3,500) against $73.9\%$ (2,585/3,500) originally, a drift of $-0.4$ pp; all comparisons in the paper use the original run.

\subsection{Answer grading}
\label{app:grading}

Responses are normalized (lowercased, whitespace-collapsed) before grading. Grading is answer-type-specific:
\begin{itemize}[nosep,leftmargin=*]
\item \textbf{yes\_no}: Response must start with the expected token (``yes'' or ``no'').
\item \textbf{number}: Expected digit must appear in the first 4 characters.
\item \textbf{left\_right}: Response must contain any of the expected aliases (e.g., ``left'', ``left box'', container color name).
\item \textbf{belief\_split} (reality-vs-belief template): The response is split at the word ``belief'' (or at midpoint if absent); the reality answer must appear before the split and the belief answer after.
\item \textbf{All other types}: Response must contain at least one expected alias.
\end{itemize}
No partial credit is awarded. The grading function is deterministic and identical across all conditions and models.

\subsection{Error categorization}
\label{app:errorcats}

Stage~3 \chipA\,Condition~A errors are classified by a deterministic rule applied to the normalized model output. A response is labeled a \textbf{reality-bias} error if it contains an ``actual alias'' (the object's veridical location or the reality-consistent answer) but does not contain the expected belief-consistent answer; all remaining errors are uncategorized. Reality-bias errors account for $549$ of $915$ visual-only Stage~3 errors across the seven-model lineup ($60.0\%$); the rule is identical across models.

\subsection{Prompt paraphrase variants}
\label{app:paraphrases}

Three paraphrases of each Stage~3 \chipA\,Condition~A question are constructed:
\begin{itemize}[nosep,leftmargin=*]
\item \textbf{Restructured}: ``This image has [$N_\mathrm{panels}$] panels showing colored figures, containers, and smaller objects. Follow the sequence carefully before answering: [task\_prompt] [answer\_instruction]''
\item \textbf{Vocabulary}: Synonym substitution applied to the original question (``character'' $\to$ ``figure'', ``container'' $\to$ ``box'', ``object'' $\to$ ``item'', ``search'' $\to$ ``look in'').
\item \textbf{Distractor}: ``The background is white. [original question]''
\end{itemize}

\section{Extended results}
\label{app:extended}

\subsection{Stages 1--2 per-model results}
\label{app:s12}

Table~\ref{tab:s12_supp} presents the complete Stages~1--2 per-model matrix showing all five conditions.

\begin{table}[t]
\centering
\caption{\textbf{Stages 1--2 per-model results} across the seven-model lineup (all five conditions, accuracy\,$\uparrow$ \%). \textbf{Bold}: best per row; \emph{italic}: 2nd.}
\label{tab:s12_supp}
\resizebox{\columnwidth}{!}{%
\scriptsize
\setlength{\tabcolsep}{3pt}
\begin{tabular}{@{}l ccccc ccccc@{}}
\toprule
& \multicolumn{5}{c}{\textbf{Stage 1: Object Tracking}} & \multicolumn{5}{c}{\textbf{Stage 2: Goal Attribution}} \\
\cmidrule(lr){2-6}\cmidrule(lr){7-11}
\textbf{Model} & \chipA\,A & \chipBa\,$B_a$ & \chipBp\,$B_p$ & \chipBf\,$B_f$ & \chipC\,C & \chipA\,A & \chipBa\,$B_a$ & \chipBp\,$B_p$ & \chipBf\,$B_f$ & \chipC\,C \\
\midrule
GPT-4o~\cite{openai2024gpt4o}      & 83.3 & 83.3 & \textbf{100} & \textbf{100} & 83.3 & 93.0 & 91.5 & \textbf{100} & \textbf{100} & \textbf{100} \\
Gemini 2.5 Pro~\cite{google2025gemini} & 71.0 & 83.3 & \emph{83.7} & 80.3 & \textbf{94.0} & 94.0 & \emph{96.0} & 94.5 & 92.5 & \textbf{100} \\
Claude Sonnet 4~\cite{anthropic2025claude} & 33.3 & 50.0 & \emph{99.3} & \emph{99.3} & \textbf{100} & 75.0 & 75.0 & \textbf{100} & \textbf{100} & 51.5$^\dagger$ \\
\midrule
Qwen2.5-VL-72B~\cite{bai2025qwen25vl} & 83.3 & 73.7 & \textbf{100} & \emph{94.7} & \textbf{100} & \textbf{100} & \textbf{100} & \textbf{100} & \textbf{100} & \textbf{100} \\
Llama 4 Mav.~\cite{meta2025llama4} & 81.3 & 80.3 & \textbf{100} & \textbf{100} & \emph{98.0} & 83.0 & 81.0 & \textbf{100} & \emph{99.5} & \textbf{100} \\
\midrule
GPT-4.1-mini~\cite{openai2025gpt41mini} & 81.7 & 82.0 & \textbf{100} & \textbf{100} & 83.3 & 91.0 & 85.0 & \textbf{100} & \textbf{100} & \textbf{100} \\
Ministral-8B~\cite{mistral2025ministral3} & 75.7 & 47.7 & \textbf{91.7} & \emph{79.3} & 67.0 & \textbf{100} & 97.5 & \textbf{100} & \textbf{100} & \textbf{100} \\
\midrule
\textbf{Mean} & 72.8 & 71.5 & \textbf{96.4} & \emph{93.4} & 89.4 & 90.9 & 89.4 & \textbf{99.2} & \emph{98.9} & 93.1 \\
\bottomrule
\multicolumn{11}{@{}l@{}}{\tiny $^\dagger$Claude text-only Stage~2 anomaly: direction-convention misinterpretation in pure-text probes (see \S\ref{app:addlimitations}).}
\end{tabular}
}
\end{table}

\subsection{Per-template Stage~3 results}
\label{app:pertemplate}

Table~\ref{tab:templates_supp} presents accuracy for all 10 Stage~3 templates.

\begin{table}[t]
\centering
\caption{\textbf{Stage~3 per-template Condition~A accuracy\,$\uparrow$ (\%)} across the seven-model lineup (350 instances per template), sorted by difficulty. Parser: structural parse accuracy (\%).}
\label{tab:templates_supp}
\footnotesize
\setlength{\tabcolsep}{4pt}
\begin{tabular}{@{}lcc@{}}
\toprule
\textbf{Template} & \chipA\,Acc\,$\uparrow$ & Parser\,$\uparrow$ \\
\midrule
\emph{unexpected\_contents} & 55.4 & 100 \\
\emph{two\_rooms} & 56.0 & 94 \\
\emph{second\_order} & 64.9 & 100 \\
\emph{reality\_vs\_belief} & 70.0 & 100 \\
\emph{knowledge\_belief} & 72.0 & 100 \\
\emph{sally\_anne\_reality} & 74.9 & 100 \\
\emph{knowledge\_who\_knows} & 80.6 & 100 \\
\emph{sally\_anne\_search} & 83.4 & 100 \\
\emph{sally\_anne\_belief} & 85.7 & 100 \\
\emph{perspectives\_differ} & 95.7 & 100 \\
\midrule
\textbf{Mean} & \textbf{73.9} & \textbf{99.4} \\
\bottomrule
\end{tabular}
\end{table}

\subsection{Prompt robustness extended}
\label{app:robustness}

The three paraphrase variants (\S\ref{app:paraphrases}) are applied to Stage~3 \chipA\,Condition~A and compared against the 7-model abstract Stage~3 baseline (73.9\%). The paraphrases are evaluated on 200 probes per model, whereas the baseline uses all 500. Per-variant accuracies over the seven-model lineup are $70.1\%$ (restructured), $71.9\%$ (vocabulary), and $69.9\%$ (distractor), within $4.0$ pp of the original (Table~\ref{tab:prompt}a). The distractor sentence differs from the adversarial insertion used for C and $B_f$. Paraphrases of $C$ and $B_f$ are reported in \S\ref{app:paraphrase_cb}; the inferential competence captured in the matched conditions is robust to phrasing.

\section{Complete template catalog}
\label{app:templates}

\begin{figure}[t]
    \centering
    \includegraphics[width=0.4\linewidth]{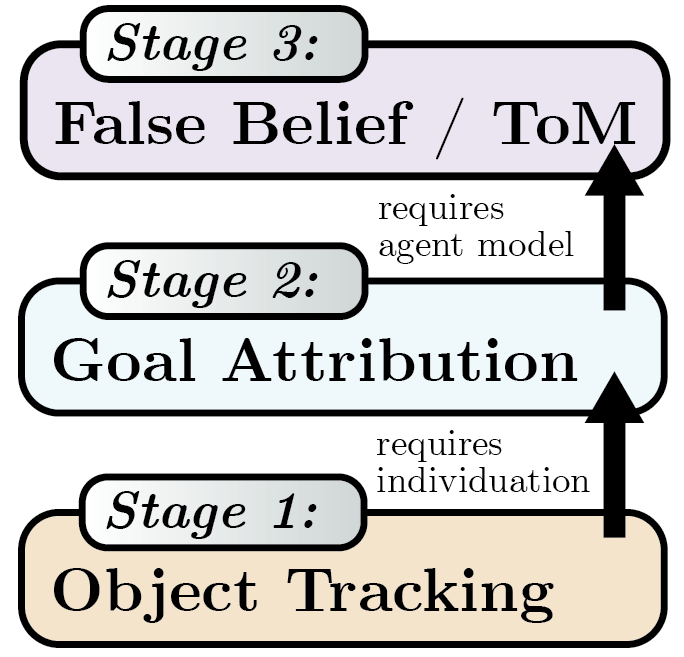}
    \caption{\textbf{Developmental stage hierarchy.} Each stage requires the representational output of the previous one, mirroring the cascade identified in human cognitive development: object individuation at 3--5 months~\cite{spelke1990principles}, goal attribution at 6--12 months~\cite{woodward1998infants,gergely1995taking}, and false-belief reasoning at ${\sim}$4 years~\cite{wimmer1983beliefs,baron1985does}. 20 templates $\times$ 50 variants = 1{,}000 probes.}
    \label{fig:stages}
\end{figure}

The 20 templates are organized into three stages mirroring the developmental cascade in human cognitive development (Figure~\ref{fig:stages}): object individuation must precede goal attribution, which must precede belief modeling. Each stage's templates are described below.

\subsection{Stage 1: Object tracking (6 templates)}

\paragraph{\emph{occlusion\_consistent} / \emph{occlusion\_violation}.}
5 panels. A colored circle moves rightward toward a gray barrier, disappears behind it (Panels 3--4), and reappears on the correct side (consistent) or the wrong side (violation). Question: ``Is the final reappearance position consistent with the earlier direction of motion, yes or no?'' Answer: yes/no.

\paragraph{\emph{count\_two} / \emph{count\_three}.}
4 panels. Two (or three) identically or distinctly colored circles follow crossing trajectories. Question: ``How many distinct circular objects are present?'' Answer: 2 or 3.

\paragraph{\emph{permanence}.}
4 panels. A colored circle enters a container (Panels 1--2), remains hidden (Panels 3--4). Question: ``In the final panel, is the circle hidden or gone?'' Answer: hidden.

\paragraph{\emph{color\_change}.}
5 panels. A colored circle moves behind a barrier and reappears in a different color. Question: ``Are the circle colors before and after the barrier the same or different?'' Answer: different.

\subsection{Stage 2: Goal attribution (4 templates)}

\paragraph{\emph{goal\_upper}.}
4 panels. An agent moves along a trajectory toward one of two targets (upper or lower). Question: ``Which target is the agent heading toward?'' Answer: upper/lower.

\paragraph{\emph{detour\_obstacle}.}
5 panels. An agent takes a curved path around a gray obstacle to reach a target. Question: ``Why did the character take the curved path instead of going straight: A) prefers curves, B) obstacle blocking, or C) avoiding the target?'' Answer: B.

\paragraph{\emph{no\_goal}.}
4 panels. A circle drifts at constant velocity without changing direction toward any target. Question: ``Is this motion goal-directed?'' Answer: no.

\paragraph{\emph{wall\_removed}.}
2 panels. Left panel: straight path with no obstacle. Right panel: curved detour path with no obstacle. Question: ``With no obstacle present, which path would a goal-directed character choose, straight or curved?'' Answer: straight.

\subsection{Stage 3: False belief / Theory of Mind (10 templates)}

\paragraph{\emph{sally\_anne\_search} / \emph{sally\_anne\_belief} / \emph{sally\_anne\_reality}.}
6 panels. Classic Sally-Anne scenario: Agent~1 places an object in Container~L, exits, Agent~2 transfers the object to Container~R, Agent~1 returns. Three question variants probe search prediction (where will Agent~1 look first? Left), belief report (where does Agent~1 think the object is? Left), and reality check (where is the object actually? Right).

\paragraph{\emph{reality\_vs\_belief}.}
6 panels. Same scenario. The model must simultaneously report both the object's actual location (right) and Agent~1's believed location (left) in a single response.

\paragraph{\emph{knowledge\_who\_knows} / \emph{knowledge\_belief}.}
5 panels. Two agents are in a room; one exits, the object is moved, the exiting agent returns. \emph{knowledge\_who\_knows}: ``Which character knows the object's current location?'' (the one who stayed). \emph{knowledge\_belief}: ``Where does the returning character think the object is?'' (left, the pre-transfer location).

\paragraph{\emph{unexpected\_contents}.}
5 panels. Agent~1 sees a box with round symbols, opens it to find rectangular contents, closes it. Agent~2 arrives and sees only the closed box with round symbols. Question: ``What does Agent~2 think is inside?'' Answer: round objects.

\paragraph{\emph{two\_rooms}.}
5 panels. Two rooms divided by a wall. Agent~1 carries an object from the left room to the right room while Agent~2 watches. Agent~2 then exits. While Agent~2 is gone, Agent~1 places the object in the right container. Agent~2 returns. Question: ``Where does Agent~2 think the object is?'' Answer: right (Agent~2 saw the transfer to the right room before leaving).

\paragraph{\emph{perspectives\_differ}.}
5 panels. Similar to Sally-Anne but the question asks whether two agents hold the same or different beliefs about the object's location. Answer: different.

\paragraph{\emph{second\_order}.}
5 panels. One agent is absent during a transfer but observed by a third party. Question: ``Does the third-party agent know that the absent agent holds a false belief?'' Answer: yes.

\section{Statistical methodology}
\label{app:statistics}

\subsection{Bootstrap confidence intervals}
\label{app:bootstrap}

All confidence intervals are computed via probe-level bootstrap resampling. For each condition, the accuracy matrix has dimensions $M \times N_\mathrm{probes}$ (models $\times$ probes). We resample $N_\mathrm{probes}$ probe indices with replacement 10,000 times, compute the mean accuracy across all models and resampled probes for each bootstrap iteration, and report the 2.5th and 97.5th percentiles as the 95\% CI.

\subsection{McNemar tests}
\label{app:mcnemar}

McNemar's test is applied to matched pairs: for each (model, probe\_id) combination, we compare whether the response was correct under condition $X$ vs.\ condition $Y$, yielding a $2 \times 2$ contingency table. The test statistic is $\chi^2 = (b - c)^2 / (b + c)$, compared against $\chi^2_1$. We apply Bonferroni correction across 12 pairwise comparisons ($\alpha_\mathrm{Bonf} = 0.05/12 = 0.0042$). Stage~3 yields $n = 7 \times 500 = 3{,}500$ paired observations per comparison. The five comparisons added for the scaffold-content ladder and corruption controls ($B_\mathrm{identity}$, $B_\mathrm{static}$, $D_\mathrm{mismatch}$, $D_\mathrm{noise}$, $D_\mathrm{shuffled}$, each against A) use the same uncorrected statistic on the same $3{,}500$ matched pairs against the original Condition~A, and are evaluated against the same $\alpha_\mathrm{Bonf} = 0.0042$. Under a single correction over all 17 comparisons ($\alpha = 0.05/17 = 0.0029$), every conclusion holds except one: $B_\mathrm{static}$ ($p = .0033$) is significant at $\alpha_\mathrm{Bonf} = 0.0042$ but not at $0.0029$, so we treat it as a weak effect.

\subsection{PCR bootstrap}
\label{app:pcrboot}

The Perceptual Contribution Ratio is bootstrapped by resampling probe indices and computing $\mathrm{PCR}_x = (\overline{\text{Acc}}_B - \overline{\text{Acc}}_A) / (\overline{\text{Acc}}_C - \overline{\text{Acc}}_A)$ for each bootstrap sample.

\subsection{Axiomatic characterization of PCR}
\label{app:pcr_axioms}

\begin{figure}[t]
    \centering
    \includegraphics[width=0.8\linewidth]{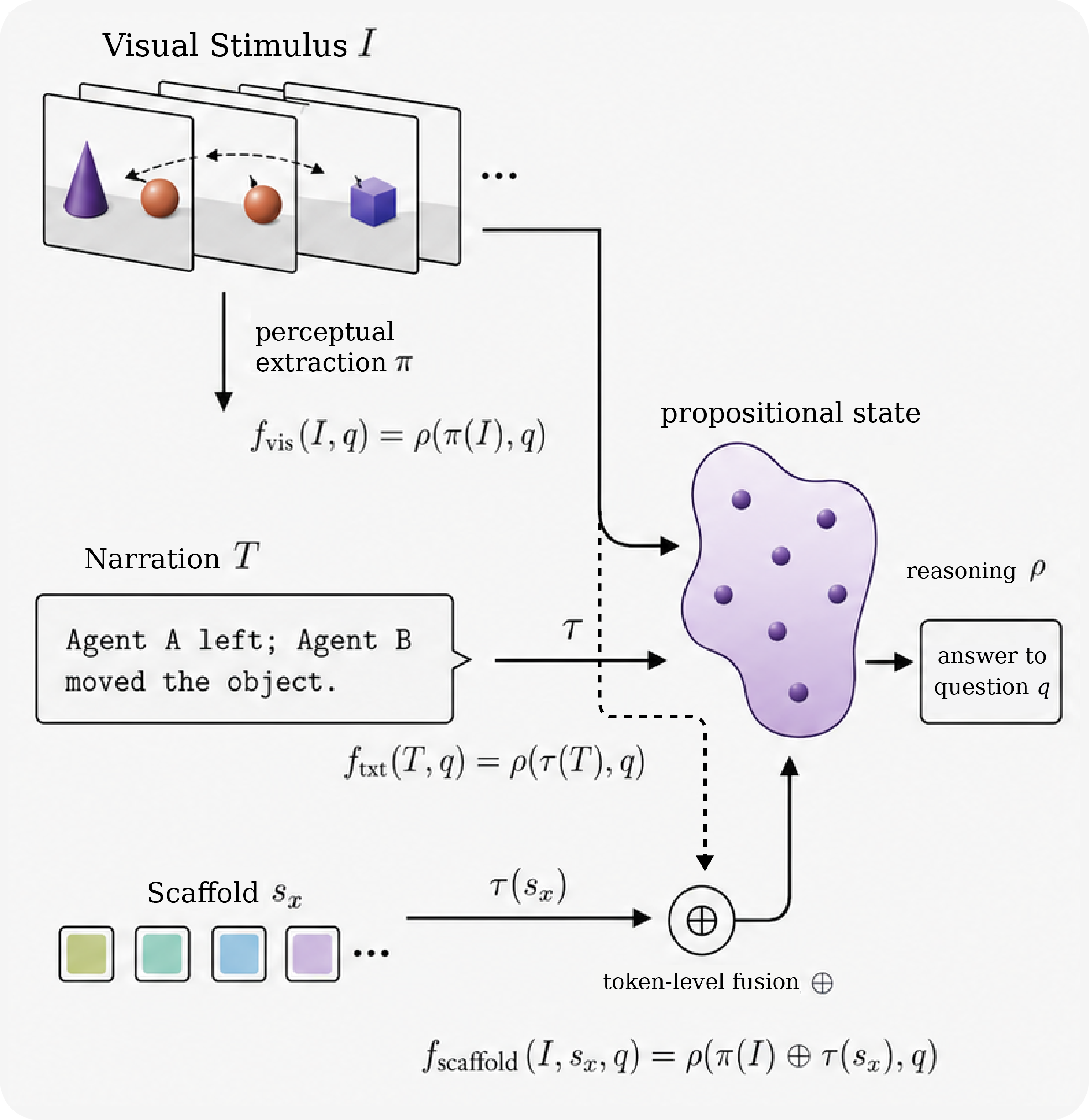}
    \caption{\textbf{Three pathways through a shared reasoning module $\rho$.} Visual input $I$ enters via perceptual extraction $\pi$ ($f_\mathrm{vis}$); a narration $T$ enters via the text encoder $\tau$ ($f_\mathrm{txt}$); a scaffold $s_x$ augments the visual path through token-level fusion $\oplus$ ($f_\mathrm{scaffold}$). The five-condition protocol (\S\ref{sec:protocol}) selects which pathway is active.}
    \label{fig:framework}
\end{figure}

The Perceptual Contribution Ratio (Definition~\ref{def:pcr}) admits a uniqueness characterization from four natural axioms. The argument formalizes PCR as the \emph{share of the closable visual-to-text gap recovered by the scaffold}, and shows that any metric respecting a small set of natural invariances must coincide with PCR.

Let $M$ be a candidate metric defined on triples $(a,b,c)$ with $c>a$, where $a=\mathrm{Acc}(A)$, $b=\mathrm{Acc}(B_x)$, and $c=\mathrm{Acc}(C)$. The pathways formalized in Figure~\ref{fig:framework} (also stated in \S\ref{sec:pcr}) supply the operational meaning: $A$ corresponds to $f_\mathrm{vis}$, $C$ to $f_\mathrm{txt}$, and $B_x$ to $f_\mathrm{scaffold}$ at fidelity $x$. Consider the following axioms.

\begin{enumerate}
\item[(A1)] \textbf{Endpoint calibration.} $M(a, a, c) = 0$ and $M(a, c, c) = 1$.
\item[(A2)] \textbf{Translation invariance.} $M(a + \delta, b + \delta, c + \delta) = M(a, b, c)$ for all $\delta \in \mathbb{R}$.
\item[(A3)] \textbf{Positive scale invariance.} $M(\alpha a, \alpha b, \alpha c) = M(a, b, c)$ for all $\alpha > 0$.
\item[(A4)] \textbf{Linearity in the scaffolded condition.} For fixed $a, c$, the map $b \mapsto M(a, b, c)$ is affine.
\end{enumerate}

Together, axioms (A2) and (A3) are equivalent to invariance under positive affine maps $z \mapsto \alpha z + \delta$ with $\alpha > 0$.

\paragraph{Interpretation.} (A1) anchors the metric to the construct: a scaffold that fails to improve over visual-only contributes nothing (purely conceptual deficit), while a scaffold that reaches the text-only ceiling has fully closed the perceptual gap. (A2) and (A3) express that accuracy is an interval-scale quantity with neither a canonical zero (chance levels differ across question formats and answer cardinalities) nor a fixed unit (proportions vs.\ percentages); the perceptual-vs-conceptual decomposition should be invariant to such reparametrizations. (A4) formalizes PCR as a \emph{share}: equal absolute gains in scaffolded accuracy contribute equally to the recovery score, and a stochastic mixture of two scaffolds receives the convex combination of their individual PCRs, the natural property for a metric reporting the fraction of a gap closed.

\begin{proposition}[Uniqueness of PCR]
\label{prop:pcr_unique}
Any function $M$ satisfying (A1) through (A4) coincides with the Perceptual Contribution Ratio:
\begin{equation}
M(a, b, c) = \frac{b - a}{c - a}.
\end{equation}
\end{proposition}

\begin{proof}
Fix $(a, b, c)$ with $c > a$. By (A2) with $\delta = -a$,
\begin{equation*}
M(a, b, c) = M(0,\, b - a,\, c - a).
\end{equation*}
Since $c > a$, the scalar $\alpha := (c - a)^{-1}$ is positive, so (A3) applies:
\begin{equation*}
M(0,\, b - a,\, c - a) = M\!\left(0,\, \tfrac{b - a}{c - a},\, 1\right) \;=:\; g\!\left(\tfrac{b-a}{c-a}\right),
\end{equation*}
where $g(t) := M(0, t, 1)$. The metric is therefore a univariate function of the affine ratio $t = (b - a)/(c - a)$. By (A1), $g(0) = 0$ and $g(1) = 1$. By (A4), $g$ is affine: $g(t) = p t + q$ for constants $p, q \in \mathbb{R}$. The endpoint conditions force $q = 0$ and $p = 1$, so $g(t) = t$ and
\begin{equation*}
M(a, b, c) = \frac{b - a}{c - a}. \qedhere
\end{equation*}
\end{proof}

\paragraph{Necessity of each axiom.} Each axiom is independently required. Dropping (A1) admits any affine rescaling $\alpha\,\mathrm{PCR} + \beta$. Dropping (A2) admits metrics that depend on the absolute origin (e.g., $b/c$). Dropping (A3) admits metrics that depend on the unit (e.g., the raw gap $b - a$). Dropping (A4) admits any endpoint-preserving reparametrization $M(a,b,c) = g\!\left((b-a)/(c-a)\right)$ with $g(0)=0$ and $g(1)=1$; on the in-range interval $t \in [0,1]$, examples include $t^k$ for $k>0$ or $\sqrt{t}$. Such reparametrizations would assign different contribution scores to the same scaffolded accuracy depending on the curvature of the chosen $g$, with no principled basis for selecting one curvature over another. Among metrics that respect all four invariances, PCR is the unique choice.

\paragraph{Out-of-range values.} The proposition makes no assumption that $b \in [a, c]$, so the characterization extends to the empirically observed regime in which $\mathrm{PCR} > 1$ (the scaffold-plus-image combination matches or exceeds the text-only ceiling; \S\ref{app:rebuttal}) and $\mathrm{PCR} < 0$ (scaffold actively interferes with reasoning, observed for the blob-detection parser on Stage~3 and for several mid-scale models on $B_a$). The same affine ratio governs all three regimes, which is what licenses the dose-response interpretation of the triplet $(\mathrm{PCR}_{a},\,\mathrm{PCR}_{p},\,\mathrm{PCR}_{f})$.

\section{Extended discussion of competing accounts}
\label{app:competing}

Several recent works argue that VLM failures on social reasoning tasks are cognitive rather than perceptual. We discuss each in detail and explain why our findings are complementary rather than contradictory.

\paragraph{Liu~\etal~\cite{liu2026seeing}: ``Seeing but Not Believing.''}
This work demonstrates that VLMs internally attend to correct visual evidence yet still output incorrect answers on VQA benchmarks. The key distinction from our work lies in the \emph{perceptual extraction complexity} of the tasks. Their benchmarks involve single-image VQA where the relevant visual evidence occupies a spatially localized region within one frame. Our stimuli require \emph{sequential multi-panel entity tracking}. Our data directly address whether this distinction holds: the 60.0\% reality-bias error rate shows that models correctly identify where the object currently is but fail to track who was absent when it moved.

\paragraph{Chen~\etal~\cite{chen2024vlms_see}: ``VLMs See What You Want but Not What You See.''}
VLMs achieve high performance on intent recognition but consistently fail at perspective-taking, scoring \emph{below chance} on PerspectBench. Our work offers a complementary interpretation: the egocentric bias they observe is precisely what our reality-bias error category captures (549 of 915 visual-only Stage~3 errors; 60.0\%). We demonstrate that this bias disappears when perceptual scaffolding provides explicit agent-perspective information (\chipBf\,$B_f$: 98.4\% vs.\ \chipA\,A: 73.9\%).

\paragraph{Tong~\etal~\cite{tong2024eyes}: ``Eyes Wide Shut.''}
Their identification of systematic visual shortcomings in multimodal LLMs traceable to CLIP-based encoders provides complementary evidence from the representation-learning side. Our scaffolding intervention demonstrates the downstream consequence: when structured perceptual content is provided externally, the reasoning succeeds, consistent with their finding that visual grounding remains an open challenge.

\paragraph{Zhang~\etal~\cite{zhang2025mindpower}: ``MindPower.''}
Their BDI framework decomposes ToM reasoning into perception, mental reasoning, decision, and action layers, finding that VLMs perform poorly across all layers but especially on mental reasoning. Our work is consistent with their per-layer analysis and adds the \emph{scaffolding intervention} that isolates the perceptual from the conceptual component.

\paragraph{Jin~\etal~\cite{jin2024mmtom}: ``MMToM-QA.''}
Our benchmark differs in three key respects: (i) Abstract 2D panels rather than naturalistic 3D game scenes; (ii) a scaffolding intervention that disentangles perception from reasoning; (iii) evaluation across a developmental hierarchy rather than testing false belief in isolation.

\paragraph{Shapira~\etal~\cite{shapira2024clever}: ``Clever Hans or Neural Theory of Mind?''}
Our work operates in the visual modality and reaches a structurally different conclusion: for frontier models, the inferential competence (demonstrated by 94.5\% \chipC\,text-only accuracy) is available in the text modality, and the visual failure is attributable to perceptual extraction, not pattern matching.

\paragraph{Hu~\etal~\cite{hu2025reevaluating}: ``Re-evaluating Theory of Mind Evaluation in LLMs.''}
Our five-condition protocol directly addresses this concern: by holding the reasoning task constant and varying only the perceptual input modality, we isolate whether the model is performing genuine belief-reality dissociation or relying on linguistic shortcuts.

\section{Additional limitations and caveats}
\label{app:addlimitations}

\paragraph{Stimulus and model generalization.}
Our evaluation covers two stimulus regimes, abstract 2D panels and naturalistic 3D game scenes from MuMA-ToM and MindPower. Generalization to in-the-wild human video, surveillance footage, or first-person egocentric video is not established and is a natural next test. The model lineup is limited to seven systems available through a commercial API during March 2026 (main protocol) and July 2026 (\S\ref{app:ladder}, \S\ref{app:corrupt}); ToM-specific fine-tunes, open-weight models with object-centric objectives, and not-yet-deployed multimodal systems are not represented.

\paragraph{Methodological scope.}
Models are accessed as black boxes through a single inference provider with deterministic decoding; we measure functional consequences on outputs, not attention patterns or internal representations. The MindPower $B_f$ scaffold is an oracle full-information condition rather than a leakage-controlled perceptual ceiling; the perceptual-only finding for MindPower rests on $B_p$, and the MindPower $\mathrm{PCR}_p$ values (Table~\ref{tab:tiers}c) are the appropriate quantitative summary on that regime. Reality-bias categorization is based on a deterministic algorithm (\S\ref{app:errorcats}) and inherits its definitions.

\paragraph{Answer balance.}
All \emph{sally\_anne\_search} (50 probes) and \emph{sally\_anne\_belief} (50 probes) have expected answer ``left.'' All \emph{sally\_anne\_reality} (50 probes) have expected answer ``right.'' Because every template contributes exactly 50 probes, template-reweighted PCR (\S\ref{app:reweighted}) coincides with the probe-level PCR.

\paragraph{Abstract 2D panels.}
The Heider-and-Simmel-style abstraction deliberately isolates cognitive capacity from appearance-based recognition. The perceptual bottleneck identified here is a lower bound: naturalistic 3D game scenes impose greater perceptual demands. Generalization from Abstract 2D panels to naturalistic 3D game scenes is tested separately by MuMA-ToM and MindPower rather than inferred from the panel experiments.

\paragraph{Perception/reasoning boundary.}
The attribution of the deficit to perception is made relative to the operational definition in \S\ref{sec:protocol}: structured scene-state extraction, including cross-frame correspondence and event detection. Cross-frame event linking is an intermediate representation that belief inference consumes, and it can itself be viewed as reasoning-relevant. The scaffold-content ladder (\S\ref{app:ladder}) shows that the recovery is driven by this event structure rather than by entity names or static positions; a definition that assigns event linking to reasoning would move that increment to the reasoning side, while leaving the finding that belief-level inference from access facts is not the limiting step.

\paragraph{Interpretation of the mismatch control.}
Because $D_\mathrm{mismatch}$ partners share the template, they carry useful template-level information while misdescribing the specific scene. Its small net effect may therefore reflect two opposing effects rather than the absence of any effect. The corruption effects are also strongly model-dependent, and $D_\mathrm{noise}$ and $D_\mathrm{shuffled}$ are weak by construction. We therefore treat the controls as evidence against a text-volume account, not as proof that the benefit requires image-bound content. Designing multimodal augmentations that jointly manipulate the visual and textual inputs while systematically varying their alignment is a promising direction for future work.

\paragraph{Broader impacts.}
This work is diagnostic: it trains no models and uses procedurally generated stimuli and existing public benchmarks, with no human subjects or personal data. On the positive side, localizing visual ToM failures to perceptual extraction can direct effort toward front-ends that make assistive, educational, and collaborative systems more reliable at tracking what the people they serve have and have not seen. On the negative side, better tracking of who saw what could also strengthen systems used for surveillance or for manipulation that exploits gaps in a person's knowledge. Our protocol measures this capability rather than providing it, and we release stimuli and evaluation code only, not models.

\paragraph{Dataset licenses.}
MuMA-ToM~\cite{shi2025muma} is released under the MIT license and MindPower~\cite{zhang2025mindpower} under Apache-2.0, as declared on their dataset cards; the simulator assets they build on may carry separate terms.

\paragraph{API-based evaluation.}
All models are accessed through the OpenRouter API as black boxes. Temperature 0 removes stochasticity, meaning each model gives one deterministic answer per probe; error bars reflect cross-probe and cross-model variance, not model uncertainty.

\paragraph{Stage~1 leading prompts.}
Three Stage~1 templates have answer instructions that contain only the expected answer: \emph{count\_two} (``Answer ONLY `2'\,''), \emph{permanence} (``Answer ONLY `hidden'\,''), and \emph{color\_change} (``Answer ONLY `different'\,''). Within-probe comparisons are unaffected, but the absolute accuracy of these templates, and hence the Stage~1 aggregates, is inflated. No Stage~3 template, which carries the paper's claims, is affected.

\paragraph{Scaffold fidelity spectrum.}
The five conditions ($A, B_a, B_p, B_f, C$) sample the scaffold fidelity continuum at discrete points. The additional $B_\mathrm{vis{-}stripped}$ and $B_\mathrm{vlm}$ conditions (\S\ref{app:rebuttal}) provide two further data points on this continuum.

\paragraph{Claude text-only Stage~2 anomaly.}
Claude Sonnet 4 scores 51.5\% on text-only goal attribution (Table~\ref{tab:s12_supp}), far below the other six models (all $100\%$). Inspection reveals a systematic direction-convention misinterpretation in pure-text probes: Claude interprets ``upper'' and ``lower'' differently from the intended spatial convention when no image is present. This anomaly is specific to Stage~2 text-only and does not affect Stage~3 conclusions.

\paragraph{Model API versions.}
All models were evaluated through OpenRouter. We report the exact model identifiers used:\\
\texttt{openai/gpt-4o-2024-11-20},\\
\texttt{google/gemini-2.5-pro-preview} (abstract regime; \texttt{google/gemini-2.5-pro} on MuMA-ToM and MindPower),\\
\texttt{anthropic/claude-sonnet-4} (served as the 2025-05-22 snapshot),\\
\texttt{qwen/qwen2.5-vl-72b-instruct},\\
\texttt{meta-llama/llama-4-maverick-17b-128e-}\\
\texttt{instruct},\\
\texttt{openai/gpt-4.1-mini-2025-04-14},\\
\texttt{mistralai/ministral-8b-2512}.


\section{Additional ablation experiments}
\label{app:rebuttal}

This section stress-tests the perceptual-bottleneck interpretation by asking seven failure-mode questions: can a stronger parser rescue the image-only condition, is the gain merely explicit visibility leakage, do text scaffolds survive paraphrase, can the model repair itself with chain-of-thought, are the results an artifact of template imbalance, which scaffold content drives the recovery, and does the benefit require correct content or merely more text? The first five ablations contribute 38,500 additional Abstract 2D panel calls; the scaffold-content ladder, the Ministral-8B corruption extension, and a re-run of Condition~A to check for API drift contribute 12,000 more; the corruption runs on the other six models are already included in the totals of \S\ref{app:api}.

\subsection{Experiment 1: VLM-as-parser (\texorpdfstring{$B_\mathrm{vlm}$}{B\_vlm})}
\label{app:vlmparser}

Two frontier VLMs are independently tested as zero-shot visual scene describers. For each of the 500 Stage~3 probes, the parser model is prompted with a system message instructing a panel-by-panel description of entity positions, colors, and inter-panel changes, without any ToM interpretation (max 600 tokens, temperature 0). The resulting narrations are then provided as scaffolds to the ablation subset reported in Table~\ref{tab:vlm_parser_supp}.

\begin{table}[t]
\centering
\caption{\textbf{\texorpdfstring{$B_\mathrm{vlm}$}{B\_vlm} per-model Stage~3 accuracy} (Stage~3 ablation subset, $n$=500). Two independent parser models.}
\label{tab:vlm_parser_supp}
\footnotesize
\setlength{\tabcolsep}{2.5pt}
\begin{tabular}{@{}l cc ccc@{}}
\toprule
\textbf{Model} & \multicolumn{2}{c}{\chipBv\,$B_\mathrm{vlm}$} & \chipA\,A & \chipBp\,$B_p$ & \chipBf\,$B_f$ \\
\cmidrule(lr){2-3}
 & GPT-4o & Claude & & & \\
 & parser & parser & & & \\
\midrule
GPT-4o~\cite{openai2024gpt4o}           & 81.6 & 77.2 & 76.4 & 90.2 & 99.2 \\
Gemini 2.5 Pro~\cite{google2025gemini}   & 79.6 & 60.8 & 71.8 & 90.4 & 92.6 \\
Claude Sonnet 4~\cite{anthropic2025claude}  & 79.8 & 76.8 & 78.8 & 97.6 & 100.0 \\
\midrule
Qwen2.5-VL-72B~\cite{bai2025qwen25vl}   & 62.0 & 59.6 & 79.2 & 90.0 & 100.0 \\
Llama 4 Mav.~\cite{meta2025llama4}      & 84.0 & 84.2 & 87.0 & 85.0 & 99.8 \\
\midrule
GPT-4.1-mini~\cite{openai2025gpt41mini}  & 72.0 & 74.8 & 70.2 & 80.2 & 100.0 \\
\midrule
\textbf{Mean}    & \textbf{76.5} & \textbf{72.2} & 77.2 & 88.9 & 98.6 \\
\textbf{PCR}     & $-$0.04 & $-$0.27 & n/a & 0.63 & 1.16 \\
\bottomrule
\end{tabular}
\end{table}

Neither parser provides an aggregate benefit over visual-only input. GPT-4o-as-parser (76.5\%) is statistically indistinguishable from Condition~A on the same six-model abstract subset (77.2\%; $\mathrm{PCR}=-0.04$); Claude-Sonnet-4-as-parser (72.2\%) degrades performance by 5.0\,pp ($\mathrm{PCR}=-0.27$). The corresponding values relative to the seven-model main aggregate (Cond~A $=73.9\%$, Cond~C $=94.5\%$) are $\mathrm{PCR}_\mathrm{vlm}^\mathrm{GPT}=0.13$ and $\mathrm{PCR}_\mathrm{vlm}^\mathrm{Claude}=-0.08$. Although both parsers generate substantially richer descriptions than the blob-detection parser, they commit systematic errors: hallucinating entities not present in panels, inconsistently naming colors across panels, and failing to detect exit events. The consistency across two independent frontier parsers confirms that the bottleneck is a fundamental limitation of current visual encoders, not an artifact of any single model's weaknesses.

\subsection{Experiment 2: Visibility ablation (\texorpdfstring{$B_\mathrm{vis{-}stripped}$}{B\_vis-stripped})}
\label{app:visstrip}

Explicit visibility/absence phrases are removed from $B_f$ scaffolds via regex matching against nine patterns, which cover the phrase families listed below. On average, $67\pm35$ characters (8.9\% of scaffold length per probe) are removed per probe, affecting 80\% of probes. Phrasings outside the pattern set (\eg ``had already seen'', ``never saw'', ``left before the object was placed'') are not removed, so the stripping is conservative and some visibility content remains in $B_\mathrm{vis{-}stripped}$.

\paragraph{Visibility patterns removed:}
\begin{itemize}[nosep,leftmargin=*]
\item Sentences containing: ``was/is absent'', ``did/does not see/observe'', ``was/is not present'', ``unaware of'', ``has no knowledge of'', ``cannot/could not see'', ``left the room/scene'', ``exited the room/scene'', ``not in the room'', ``outside the room'', ``while X was gone/away'', ``without X knowing/seeing''.
\end{itemize}

\begin{table}[t]
\centering
\caption{\textbf{\texorpdfstring{$B_\mathrm{vis{-}stripped}$}{B\_vis-stripped} per-model Stage~3 accuracy} ($n$=500), six models with logged runs (Ministral-8B was not evaluated in this condition). Compared against $B_p$ and $B_f$ on the same six models.}
\label{tab:vis_stripped_supp}
\footnotesize
\setlength{\tabcolsep}{3pt}
\begin{tabular}{@{}l c ccc@{}}
\toprule
\textbf{Model} & \chipBs\,$B_\mathrm{vis{-}stripped}$ & \chipBp\,$B_p$ & \chipBf\,$B_f$ & $\Delta$ vs $B_f$ \\
\midrule
GPT-4o~\cite{openai2024gpt4o}           & 99.6 & 90.2 & 99.2 & \up{+0.4} \\
Gemini 2.5 Pro~\cite{google2025gemini}   & 54.4 & 90.4 & 92.6 & \dn{$-$38.2} \\
Claude Sonnet 4~\cite{anthropic2025claude}  & 100.0 & 97.6 & 100.0 & 0.0 \\
\midrule
Qwen2.5-VL-72B~\cite{bai2025qwen25vl}   & 99.8 & 90.0 & 100.0 & \dn{$-$0.2} \\
Llama 4 Mav.~\cite{meta2025llama4}      & 98.2 & 85.0 & 99.8 & \dn{$-$1.6} \\
\midrule
GPT-4.1-mini~\cite{openai2025gpt41mini}  & 88.0 & 80.2 & 100.0 & \dn{$-$12.0} \\
\midrule
\textbf{Mean}    & \textbf{90.0} & 88.9 & 98.6 & \dn{$-$8.6} \\
\bottomrule
\end{tabular}
\end{table}

The aggregate result over the six evaluated models ($B_\mathrm{vis{-}stripped}=90.0\%$; $\mathrm{PCR}_\mathrm{vis{-}stripped}=0.69$, computed against A $=77.2\%$ and C $=95.7\%$ on the same models) confirms that most of the scaffold benefit derives from perceptual content rather than explicit visibility predicates. Gemini 2.5 Pro is a notable outlier (54.4\%), suggesting model-specific reliance on explicit absence cues; all other models remain at or above 88.0\%.

\subsection{Experiment 3: Paraphrase robustness for C and \texorpdfstring{$B_f$}{B\_f}}
\label{app:paraphrase_cb}

Three paraphrase variants are applied to both Condition~C and $B_f$ scaffolds across the ablation subset and 500 probes:
\begin{itemize}[nosep,leftmargin=*]
\item \textbf{Restructured}: Panel sentences in reverse order.
\item \textbf{Vocabulary}: Synonym substitution (``character''$\to$``figure'', ``container''$\to$``box'', etc.).
\item \textbf{Adversarial}: Ullman-style misleading insertion: ``Note that all characters could potentially see everything that happens in the scene.''
\end{itemize}

\begin{table}[t]
\centering
\caption{\textbf{Paraphrase robustness for C and $B_f$} (Stage~3 ablation subset, $n$=3,000 per variant).}
\label{tab:paraphrase_supp}
\footnotesize
\setlength{\tabcolsep}{3pt}
\begin{tabular}{@{}lcc@{}}
\toprule
\textbf{Condition} & \textbf{Acc (\%)} & $\Delta$ vs original \\
\midrule
C (original) & 95.7 & n/a \\
$C_\mathrm{restructured}$ & 95.2 & \dn{$-$0.5} \\
$C_\mathrm{vocabulary}$ & 94.8 & \dn{$-$0.9} \\
$C_\mathrm{adversarial}$ & 97.2 & \up{+1.5} \\
\midrule
$B_f$ (original) & 98.6 & n/a \\
$B_{f,\mathrm{restructured}}$ & 93.9 & \dn{$-$4.7} \\
$B_{f,\mathrm{vocabulary}}$ & 97.1 & \dn{$-$1.5} \\
$B_{f,\mathrm{adversarial}}$ & 97.6 & \dn{$-$1.0} \\
\bottomrule
\end{tabular}
\end{table}

All six paraphrase conditions exceed 93\%, confirming that the inferential competence measured by these conditions is robust and not phrasing-dependent.

\subsection{Experiment 4: Chain-of-thought self-scaffolding}
\label{app:cot}

Each model in this ablation receives a system prompt instructing step-by-step panel description, presence/absence tracking, and then answering (max 600 tokens, temperature 0). The CoT prompt adds no external information.

\begin{table}[t]
\centering
\caption{\textbf{Chain-of-thought self-scaffolding} (Stage~3 ablation subset, $n$=500 per model). $\dagger$Gemini produces truncated markdown reasoning; only 15\% of responses were parseable.}
\label{tab:cot_supp}
\footnotesize
\setlength{\tabcolsep}{3pt}
\begin{tabular}{@{}l cc c@{}}
\toprule
\textbf{Model} & \chipAv\,$A_\mathrm{CoT}$ & \chipA\,A & $\Delta$ \\
\midrule
GPT-4o~\cite{openai2024gpt4o}           & 76.6 & 76.4 & \up{+0.2} \\
Gemini 2.5 Pro$^\dagger$~\cite{google2025gemini}   & N/A & 71.8 & N/A \\
Claude Sonnet 4~\cite{anthropic2025claude}  & 90.6 & 78.8 & \up{+11.8} \\
\midrule
Qwen2.5-VL-72B~\cite{bai2025qwen25vl}   & 74.0 & 79.2 & \dn{$-$5.2} \\
Llama 4 Mav.~\cite{meta2025llama4}      & 62.8 & 87.0 & \dn{$-$24.2} \\
\midrule
GPT-4.1-mini~\cite{openai2025gpt41mini}  & 79.0 & 70.2 & \up{+8.8} \\
\midrule
\textbf{Mean (excl.\ Gem.)}    & \textbf{76.6} & 78.3 & \dn{$-$1.7} \\
\bottomrule
\end{tabular}
\end{table}

Chain-of-thought helps Claude Sonnet~4 ($+11.8$ pp) and GPT-4.1-mini ($+8.8$ pp), is neutral for GPT-4o ($+0.2$ pp), and hurts Qwen2.5-VL-72B ($-5.2$ pp) and Llama~4 Maverick ($-24.2$ pp); the mean over the five models with parseable outputs is $-1.7$ pp. Elicited step-by-step state tracking can therefore partially compensate where a model has marginal extraction capacity, but for no model does it reach the level of supplied scene state: Claude's chain-of-thought score ($90.6\%$) remains below its own $B_p$ ($97.6\%$). This is consistent with perception and reasoning being intertwined at the implementation level. Our claim is functional rather than mechanistic: whatever the internal entanglement, the deficit is closed by supplying scene state and is not closed by prompting for more inference.

\subsection{Experiment 5: Self-scaffold generate-then-reason}
\label{app:selfscaff}

Each model in this ablation first generates a panel-by-panel scene description (max 600 tokens, temperature 0, no access to the question), then reasons over its own description together with the image (max 200 tokens, temperature 0). Two API calls are used per probe.

\begin{table}[t]
\centering
\caption{\textbf{Self-scaffold generate-then-reason baseline} (Stage~3 ablation subset on Abstract 2D panels). $A_\mathrm{self}$ asks the model to first describe the panels, then answer using its own description.}
\label{tab:selfscaff_supp}
\footnotesize
\setlength{\tabcolsep}{3pt}
\begin{tabular}{@{}l ccc@{}}
\toprule
\textbf{Model} & \chipAv\,$A_\mathrm{self}$ & \chipA\,A & $\Delta$ \\
\midrule
GPT-4o~\cite{openai2024gpt4o}           & 77.6 & 76.4 & \up{+1.2} \\
Gemini 2.5 Pro~\cite{google2025gemini}   & 14.6 & 71.8 & \dn{$-$57.2} \\
Claude Sonnet 4~\cite{anthropic2025claude}  & 69.2 & 78.8 & \dn{$-$9.6} \\
\midrule
Qwen2.5-VL-72B~\cite{bai2025qwen25vl}   & 58.6 & 79.2 & \dn{$-$20.6} \\
Llama 4 Mav.~\cite{meta2025llama4}      & 56.2 & 87.0 & \dn{$-$30.8} \\
\midrule
GPT-4.1-mini~\cite{openai2025gpt41mini}  & 74.6 & 70.2 & \up{+4.4} \\
\midrule
\textbf{Mean}    & \textbf{58.5} & 77.2 & \dn{$-$18.7} \\
\bottomrule
\end{tabular}
\end{table}

Self-scaffolding ($A_\mathrm{self}$) yields lower aggregate accuracy than standard Condition~A. Only GPT-4o ($+$1.2\,pp) and GPT-4.1-mini ($+$4.4\,pp) show marginal gains; Gemini collapses under the generate-then-reason format, and the other affected models lose accuracy. The result is consistent with the $B_\mathrm{vlm}$ finding (\S\ref{app:vlmparser}): models' own visual descriptions contain too many entity-identity and event-detection errors to serve as effective scaffolds for downstream belief reasoning.

\subsection{Experiment 6: Scaffold-content ladder (\texorpdfstring{$B_\mathrm{identity}$, $B_\mathrm{static}$}{B\_identity, B\_static})}
\label{app:ladder}

To separate perceptual repair from offloaded state tracking, two rungs are added below $B_p$ and evaluated on the same 500 Stage~3 probes and the full seven-model lineup, with the image and question unchanged. $B_\mathrm{identity}$ lists the entities in each panel, without positions; because the lists are per panel, an entity's absence from a panel is implicit in its omission. $B_\mathrm{static}$ adds per-panel entity positions and presence but no events, no cross-panel linkage, and no explicit visibility or absence statements. Both are generated programmatically from the scene generator's state. Every $B_\mathrm{static}$ scaffold was validated automatically against a 33-entry forbidden-term list using case-insensitive substring matching, with zero hits across all 500 scaffolds: \emph{belief, believe, intention, goal, did not see, didn't see, was outside, was absent, moves, moved, places, placed, exits, exited, returns, returned, enters, entered, carries, carried, transfers, transferred, opens, opened, closes, closed, sees, saw, looking, looks, searches, search, knows}. Example scaffolds for \texttt{s3\_two\_rooms\_000}, the probe of Figure~\ref{fig:qualitative}b:
\begin{itemize}[nosep,leftmargin=*]
\item $B_\mathrm{identity}$: \emph{Panel 1: red character; blue character; cyan object; mint container. Panel 2: red character; blue character; cyan object. Panel 3: red character; blue character; cyan object. Panel 4: red character; cyan object; brown container. Panel 5: blue character; cyan object; brown container.}
\item $B_\mathrm{static}$: \emph{Panel 1: red character is in the left room; cyan object is near the left mint container; blue character is in the right room. Panel 2: red character is present in the room; cyan object is present; blue character is present in the room. Panel 3: red character is in the right room; cyan object is in the right room. Panel 4: red character is present in the room; cyan object is in the right brown container. Panel 5: blue character is present in the room; cyan object is in the right brown container.}
\end{itemize}

\begin{table}[t]
\centering
\caption{\textbf{Scaffold-content ladder per model} on Stage~3 ($n$=500 per model). $B_\mathrm{identity}$: entities present in each panel, no positions. $B_\mathrm{static}$: per-panel positions, no events, linkage, or visibility.}
\label{tab:ladder_supp}
\footnotesize
\setlength{\tabcolsep}{3pt}
\begin{tabular}{@{}l ccccc@{}}
\toprule
\textbf{Model} & \chipA\,A & $B_\mathrm{identity}$ & $B_\mathrm{static}$ & \chipBp\,$B_p$ & \chipBf\,$B_f$ \\
\midrule
GPT-4o~\cite{openai2024gpt4o} & 76.4 & 79.4 & 88.2 & 90.2 & 99.2 \\
Gemini 2.5 Pro~\cite{google2025gemini} & 71.8 & 80.0 & 78.4 & 90.4 & 92.6 \\
Claude Sonnet 4~\cite{anthropic2025claude} & 78.8 & 67.6 & 71.8 & 97.6 & 100.0 \\
\midrule
Qwen2.5-VL-72B~\cite{bai2025qwen25vl} & 79.2 & 67.6 & 50.6 & 90.0 & 100.0 \\
Llama 4 Mav.~\cite{meta2025llama4} & 87.0 & 77.2 & 83.0 & 85.0 & 99.8 \\
\midrule
GPT-4.1-mini~\cite{openai2025gpt41mini} & 70.2 & 73.8 & 83.4 & 80.2 & 100.0 \\
Ministral-8B~\cite{mistral2025ministral3} & 53.6 & 72.6 & 81.2 & 88.4 & 97.4 \\
\midrule
\textbf{Mean} & 73.9 & 74.0 & 76.7 & 88.8 & \textbf{98.4} \\
\bottomrule
\end{tabular}
\end{table}

On aggregate, per-panel entity lists and static positions contribute little: $B_\mathrm{identity}$ is statistically indistinguishable from A ($+0.1$ pp, $p = .84$), even though its per-panel lists already imply who is absent from which panel, and $B_\mathrm{static}$ is small but significant at $\alpha_\mathrm{Bonf}$ ($+2.8$ pp, $p = .0033$; \S\ref{app:mcnemar}). The single largest increment ($+12.1$ pp, $B_\mathrm{static} \to B_p$) comes from cross-frame event structure, and the remaining epistemic increment $\mathrm{Acc}(B_f) - \mathrm{Acc}(B_p) = 9.6$ pp consists mainly of third-person access facts (``purple character was outside when the object moved''); the second-order template's $B_f$ additionally states the resulting first-order belief. The aggregate hides strong per-model heterogeneity (Table~\ref{tab:ladder_supp}). Static positions alone recover most or all of the $B_p$ gain for GPT-4o ($+11.8$ of $+13.8$ pp), GPT-4.1-mini ($+13.2$ \vs $+10.0$), and Ministral-8B ($+27.6$ of $+34.8$). They hurt Qwen2.5-VL-72B ($-28.6$), Claude Sonnet~4 ($-7.0$), and Llama~4 Maverick ($-4.0$). Entity lists alone hurt three models (Claude Sonnet~4, Qwen2.5-VL-72B, Llama~4 Maverick). The cross-frame event increment is therefore an aggregate statement, driven most by the models that cannot use static positions.

\paragraph{Transcript analysis.} We inspected the first $50$ incorrect saved $B_\mathrm{static}$ responses per model, in file order, for each of two models; failure labels were assigned by simple response-text rules. For Qwen2.5-VL-72B, $47$ of $50$ remain exactly the reality-bias signature (\S\ref{app:errorcats}), showing that per-panel state without event structure leaves the signature failure intact; the missing input is the event and correspondence structure that $B_p$ supplies. For Claude Sonnet~4, all $50$ inspected failures ended at the fixed 200-token budget (\texttt{finish\_reason = length}), and $48$ were overlong or truncated before reaching the required categorical answer, so its $B_\mathrm{static}$ accuracy mainly reflects truncation rather than a reasoning error.

\subsection{Experiment 7: Length-matched corrupted scaffolds}
\label{app:corrupt}

Three corruptions of the correct perceptual scaffold $B_p$ test whether gains require correct content rather than scaffold format or text volume. Each keeps the format of $B_p$ and is evaluated on the same 500 Stage~3 probes and seven models.
\begin{itemize}[nosep,leftmargin=*]
\item $D_\mathrm{mismatch}$: the $B_p$ scaffold of a different variant of the same template, with whitespace-token length matched exactly. Within each template, variants are sorted and each is paired with the next, wrapping around, so all 500 partners share their target's expected answer.
\item $D_\mathrm{noise}$: the correct $B_p$ scaffold in which each entity name is independently replaced, with probability $0.3$, by a different name from its category. This is not a one-to-one permutation; collisions and cascading substitutions can occur, and most names remain correct.
\item $D_\mathrm{shuffled}$: the correct $B_p$ scaffold with panel sentences reordered. The ``Panel~$k$'' prefixes are retained, so the original order is recoverable; we treat it as the weakest corruption.
\end{itemize}

\begin{table}[t]
\centering
\caption{\textbf{Length-matched corrupted scaffolds per model} on Stage~3 ($n$=500 per model). All corruptions are derived from $B_p$. Six-model means without Ministral-8B: A 77.2, $D_\mathrm{mismatch}$ 75.3, $D_\mathrm{noise}$ 79.4, $D_\mathrm{shuffled}$ 79.6.}
\label{tab:corrupt_supp}
\footnotesize
\setlength{\tabcolsep}{3pt}
\begin{tabular}{@{}l ccccc@{}}
\toprule
\textbf{Model} & \chipA\,A & $D_\mathrm{mismatch}$ & $D_\mathrm{noise}$ & $D_\mathrm{shuffled}$ & \chipBp\,$B_p$ \\
\midrule
GPT-4o~\cite{openai2024gpt4o} & 76.4 & 87.4 & 88.6 & 89.6 & 90.2 \\
Gemini 2.5 Pro~\cite{google2025gemini} & 71.8 & 45.0 & 53.8 & 57.6 & 90.4 \\
Claude Sonnet 4~\cite{anthropic2025claude} & 78.8 & 79.6 & 85.0 & 77.6 & 97.6 \\
\midrule
Qwen2.5-VL-72B~\cite{bai2025qwen25vl} & 79.2 & 83.0 & 87.6 & 87.2 & 90.0 \\
Llama 4 Mav.~\cite{meta2025llama4} & 87.0 & 84.2 & 82.0 & 86.4 & 85.0 \\
\midrule
GPT-4.1-mini~\cite{openai2025gpt41mini} & 70.2 & 72.6 & 79.6 & 79.0 & 80.2 \\
Ministral-8B~\cite{mistral2025ministral3} & 53.6 & 71.0 & 79.6 & 86.6 & 88.4 \\
\midrule
\textbf{Mean} & 73.9 & 74.7 & 79.5 & 80.6 & \textbf{88.8} \\
$\Delta$ \vs A & -- & \up{+0.8} & \up{+5.6} & \up{+6.7} & \up{+14.9} \\
\bottomrule
\end{tabular}
\end{table}

On aggregate, all three corruptions recover less than half of the $14.9$ pp gain that the correct scaffold adds over A, while $B_p$ itself differs from A at $p \approx 4 \times 10^{-63}$ (McNemar, 3,500 matched pairs, \S\ref{app:mcnemar}). $D_\mathrm{mismatch}$ gives no significant gain ($+0.8$ pp; $b = 520$, $c = 549$, $p = .38$). $D_\mathrm{noise}$ ($+5.6$ pp; $b = 422$, $c = 618$, $p = 1.2 \times 10^{-9}$) and $D_\mathrm{shuffled}$ ($+6.7$ pp; $b = 413$, $c = 648$, $p = 5.4 \times 10^{-13}$) give small but significant gains, well below the $+14.9$ pp of $B_p$ ($b = 226$, $c = 750$). These gains come mainly from Ministral-8B ($+17.4$ to $+33.0$ pp over its A); on the other six models the effects are $-1.9$, $+2.2$, and $+2.3$ pp (A $= 77.2\%$).

The per-model pattern is heterogeneous. Gemini 2.5 Pro falls well below A under every corruption ($-14.2$ to $-26.8$ pp) and Claude Sonnet~4 retains little of its $B_p$ gain, whereas GPT-4o retains nearly all of it ($+11.0$ to $+13.2$ of $+13.8$ pp), and GPT-4.1-mini and Ministral-8B retain most of it under $D_\mathrm{noise}$ and $D_\mathrm{shuffled}$. Because every $D_\mathrm{mismatch}$ partner shares the template and expected answer, it preserves the template's event structure while misdescribing the scene; the models that retain the gain under it appear to read that event structure from the text. The controls thus argue against text volume as the source of the gains, but they do not show that the benefit requires content bound to the specific image (\S\ref{app:addlimitations}).

\subsection{Scaffold length}
\label{app:length}

Scaffold length does not track accuracy. The zero-shot VLM-parser scaffolds (\S\ref{app:vlmparser}) are the longest we use, with a mean of 343 whitespace tokens, yet give no improvement over A ($\mathrm{PCR} = -0.04$ and $-0.27$ for the two parsers). The structured scaffolds span 107 to 136 tokens, and the text-only prompt C, which yields the strongest text-side accuracy, is the shortest at about 40 tokens. Finally, image-plus-scaffold exceeds text-only ($B_f$ $98.4\%$ \vs C $94.5\%$, $\mathrm{PCR}_f > 1$), but because the $B_f$ scaffold and the C narration differ in their text as well as in the image, this comparison does not by itself isolate the contribution of the image. On MindPower, where C is exactly the $B_p$ description without the storyboard, adding the storyboard lowers accuracy ($B_p$ $24.1\%$ \vs C $33.0\%$).

\subsection{MuMA-ToM scaffold audit}
\label{app:muma_audit}

We audited all 900 logged MuMA-ToM $B_f$ prompts against the nine visibility-pattern families of \S\ref{app:visstrip}, refactored into equivalent regular expressions. The scaffolds contain entity identities, locations, and actions, and none matches any pattern (0 of 900). This establishes the absence of regex matches, not of every possible visibility cue. Within that limit, the MuMA-ToM visual-to-text gap ($\mathrm{PCR}_f = 1.44$; frontier $1.81$) is closed and exceeded by scaffolds without explicit visibility statements.

\subsection{Template-reweighted PCR}
\label{app:reweighted}

To address potential inflation from answer-distribution imbalance, we compute PCR after reweighting: each of the 10 Stage~3 templates contributes equally to accuracy regardless of its probe count. Because every template contributes exactly 50 probes in our design, template-reweighting is equivalent to probe-averaging here, so the reweighted values coincide with the main aggregates: $A=73.9\%$, $B_a=75.7\%$, $B_p=88.8\%$, $B_f=98.4\%$, $C=94.5\%$, with $\mathrm{PCR}_a^\mathrm{rew}=0.09$, $\mathrm{PCR}_p^\mathrm{rew}=0.73$, $\mathrm{PCR}_f^\mathrm{rew}=1.19$. The same qualitative pattern holds: automatic scaffolding is weak, perceptual scaffolding recovers a large fraction of the deficit, and full scaffolding closes and slightly exceeds the text-only gap.

\clearpage
\section{Abstract 2D panels dataset overview}
\label{app:dataset_overview}

\raggedbottom
\graphicspath{{figures/all_templates/}{sampleofsyn/figures/}{figures/panels/}{figures/}}

\makeatletter
\@ifundefined{LTleft}{\newlength{\LTleft}}{}
\@ifundefined{LTright}{\newlength{\LTright}}{}
\@ifundefined{longtable}{%
  \newenvironment{longtable}[1]{\begin{center}\begin{tabular}{#1}}{\end{tabular}\end{center}}
  \@ifundefined{endfirsthead}{\def\endfirsthead{}}{}%
  \@ifundefined{endhead}{\def\endhead{}}{}%
}{}
\makeatother

\newcolumntype{L}[1]{>{\raggedright\arraybackslash}p{#1}}
\newcolumntype{C}[1]{>{\centering\arraybackslash}p{#1}}

\newcommand{\exampledivider}{%
\par\medskip
\noindent\rule{\linewidth}{0.5pt}
\par
\medskip
}

\newcommand{\textonlybox}[1]{%
  \begin{tikzpicture}
    \node[draw=condC!70!black, fill=condC!40, line width=1pt,
          rounded corners=5pt, inner sep=8pt] (tbox) {%
      \begin{minipage}{0.92\linewidth}
        {\small\ttfamily\raggedright\detokenize{#1}\par}
      \end{minipage}};
  \end{tikzpicture}%
}

\newcommand{\caseheader}[4]{%
\subsection*{\texttt{\detokenize{#1}} (Stage #2)}
\textbf{\chipC\,Text-only scene description}\par
\smallskip
\begin{center}
\textonlybox{#4}
\end{center}
\medskip
\textbf{Visual panels}\par
\smallskip
\begin{center}
\stagepanel{%
\includegraphics[width=0.92\linewidth]{figures/all_templates/#3}%
}
\end{center}
}

\newcommand{\outputrow}[4]{%
\texttt{\detokenize{#1}} & #2 & #3 & \texttt{\detokenize{#4}} \\
}

\newcommand{\gridpanel}[2]{%
\begin{minipage}[t]{0.46\linewidth}
\centering
\includegraphics[width=\linewidth]{figures/all_templates/#1}\\[-2pt]
{\scriptsize\texttt{\detokenize{#2}}}
\end{minipage}
}

\newcommand{\gridpaneltwopanel}[2]{%
\begin{minipage}[t]{0.46\linewidth}
\centering
\includegraphics[width=0.78\linewidth]{figures/all_templates/#1}\\[-2pt]
{\scriptsize\texttt{\detokenize{#2}}}
\end{minipage}
}

The Abstract 2D panels benchmark is designed to evaluate whether VLM failures on visual
false-belief tasks are better explained by perceptual bottlenecks or by
reasoning bottlenecks. It spans object tracking,
goal attribution, and Theory of Mind cases, and it is intentionally organized
to mirror the developmental cascade described in the paper: object
individuation at 3--5 months, goal attribution at 6--12 months, and
false-belief reasoning at roughly 4 years. This developmental structure is
part of the benchmark's justification: later false-belief tasks depend on the
representational outputs of earlier perceptual achievements, so failures on
visual ToM probes may reflect missing perceptual state extraction rather than
missing belief reasoning. The benchmark is also intentionally designed to keep
the answer space short and tightly controlled so that errors are less
confounded by free-form generation. Most targets are therefore simple outputs
such as \texttt{yes/no}, \texttt{left/right}, a number, or a short phrase,
with only two genuinely multi-token templates:
\texttt{unexpected\_contents} (\texttt{round objects}) and
\texttt{reality\_vs\_belief}
(\texttt{Actually: right. Belief: left.}). The examples below focus on
representative cases that make the contrast between image-based and text-only
conditions especially easy to inspect.

\subsection*{1. Question inventory}

\subsection*{Stage 1: Object tracking}
\begin{center}
\footnotesize
\setlength{\tabcolsep}{3pt}
\renewcommand{\arraystretch}{1.2}
\begin{tabular}{@{}L{0.24\linewidth}L{0.43\linewidth}C{0.16\linewidth}C{0.10\linewidth}@{}}
\toprule
\textbf{Template} & \textbf{Question family} & \textbf{Answer format} & \textbf{Example} \\
\midrule
\texttt{occlusion\_consistent} & Motion continuity under occlusion & yes/no & yes \\
\texttt{occlusion\_violation} & Motion-continuity violation under occlusion & yes/no & no \\
\texttt{count\_two} & Object counting across panels & number & 2 \\
\texttt{count\_three} & Object counting across panels & number & 3 \\
\texttt{permanence} & Hidden-versus-gone judgment after occlusion & one word & hidden \\
\texttt{color\_change} & Identity or color continuity across occlusion & one word & different \\
\bottomrule
\end{tabular}
\end{center}

\subsection*{Stage 2: Goal attribution}
\begin{center}
\footnotesize
\setlength{\tabcolsep}{3pt}
\renewcommand{\arraystretch}{1.2}
\begin{tabular}{@{}L{0.24\linewidth}L{0.43\linewidth}C{0.16\linewidth}C{0.10\linewidth}@{}}
\toprule
\textbf{Template} & \textbf{Question family} & \textbf{Answer format} & \textbf{Example} \\
\midrule
\texttt{goal\_upper} & Goal direction toward one of two targets & upper/lower & upper \\
\texttt{detour\_obstacle} & Cause of a detour around an obstacle & A/B/C choice & B \\
\texttt{no\_goal} & No-goal control: movement not directed to a target & yes/no & no \\
\texttt{wall\_removed} & Rational path choice when the obstacle is removed & straight/curved & straight \\
\bottomrule
\end{tabular}
\end{center}

\subsection*{Stage 3: False belief / Theory of Mind}
\begin{center}
\footnotesize
\setlength{\tabcolsep}{3pt}
\renewcommand{\arraystretch}{1.2}
\begin{tabular}{@{}L{0.24\linewidth}L{0.43\linewidth}C{0.16\linewidth}C{0.10\linewidth}@{}}
\toprule
\textbf{Template} & \textbf{Question family} & \textbf{Answer format} & \textbf{Example} \\
\midrule
\texttt{sally\_anne\_search} & False-belief search prediction & left/right & left \\
\texttt{sally\_anne\_belief} & False-belief location attribution & left/right & left \\
\texttt{sally\_anne\_reality} & Reality-location check & left/right & right \\
\texttt{reality\_vs\_belief} & Joint reality-versus-belief dissociation & two-slot & Actual: right; belief: left \\
\texttt{knowledge\_who\_knows} & Knowledge attribution & stayed/left & stayed \\
\texttt{knowledge\_belief} & Belief after an unseen transfer & left/right & left \\
\texttt{unexpected\_contents} & Misleading-container contents belief & multi-token & round objects \\
\texttt{two\_rooms} & Cross-room partial observability & left/right & right \\
\texttt{perspectives\_differ} & Same-versus-different belief comparison & same/different & different \\
\texttt{second\_order} & Second-order belief or knowledge & yes/no & yes \\
\bottomrule
\end{tabular}
\end{center}

\subsection*{2. Visual template grid}

\textbf{Representative strips.} The grid below shows one saved panel sequence
for each template, organized by developmental stage.

\vspace{-10pt}
\begin{center}
\small

\textbf{Stage 1: Object Tracking}\\[3pt]
\stagepanel{%
\setlength{\tabcolsep}{2pt}\renewcommand{\arraystretch}{1.0}
\begin{tabular}{@{}cc@{}}
\gridpanel{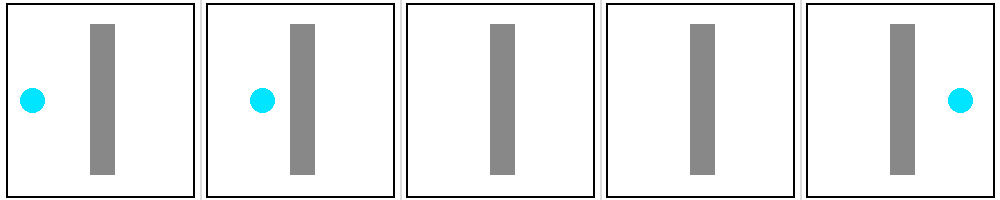}{occlusion\_consistent} &
\gridpanel{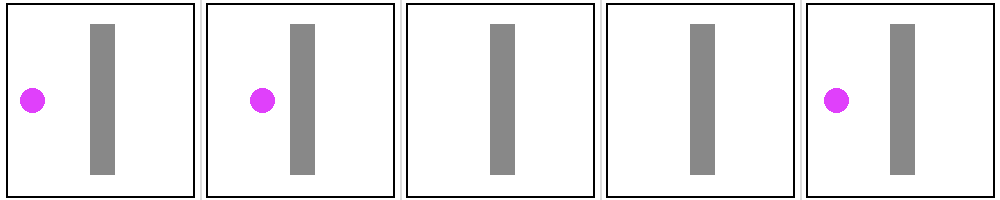}{occlusion\_violation} \\[6pt]
\gridpanel{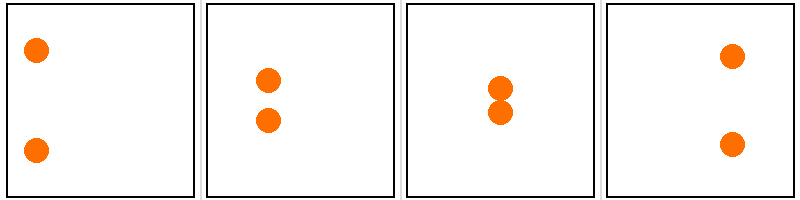}{count\_two} &
\gridpanel{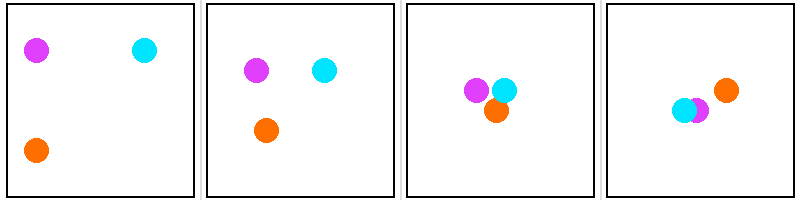}{count\_three} \\[6pt]
\gridpanel{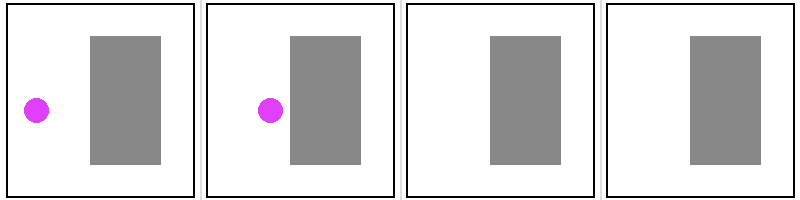}{permanence} &
\gridpanel{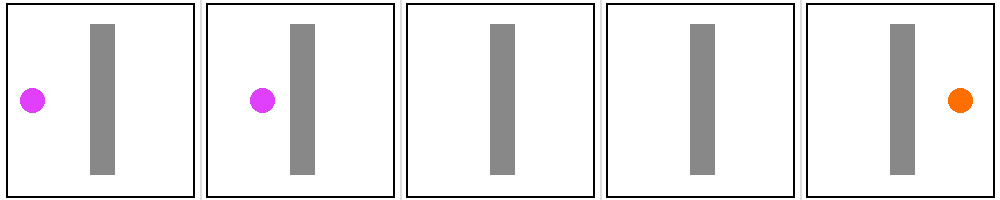}{color\_change} \\
\end{tabular}}

\vspace{-5pt}
\medskip
\textbf{Stage 2: Goal Attribution}\\[3pt]
\stagepanel{%
\setlength{\tabcolsep}{2pt}\renewcommand{\arraystretch}{1.0}
\begin{tabular}{@{}cc@{}}
\gridpanel{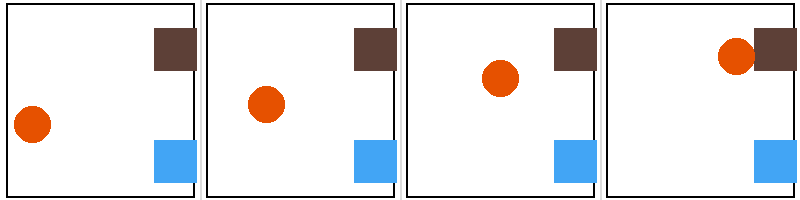}{goal\_upper} &
\gridpanel{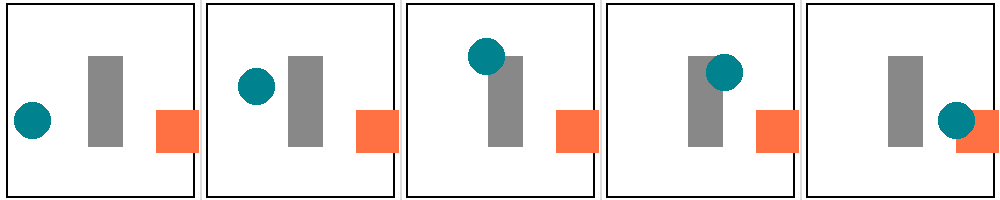}{detour\_obstacle} \\[6pt]
\gridpanel{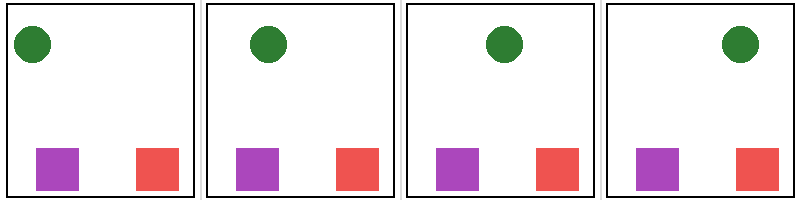}{no\_goal} &
\gridpaneltwopanel{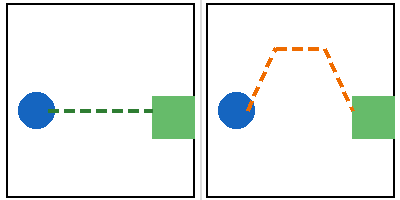}{wall\_removed} \\
\end{tabular}}

\vspace{-5pt}
\medskip
\textbf{Stage 3: False Belief / Theory of Mind}\\[3pt]
\stagepanel{%
\setlength{\tabcolsep}{2pt}\renewcommand{\arraystretch}{1.0}
\begin{tabular}{@{}cc@{}}
\gridpanel{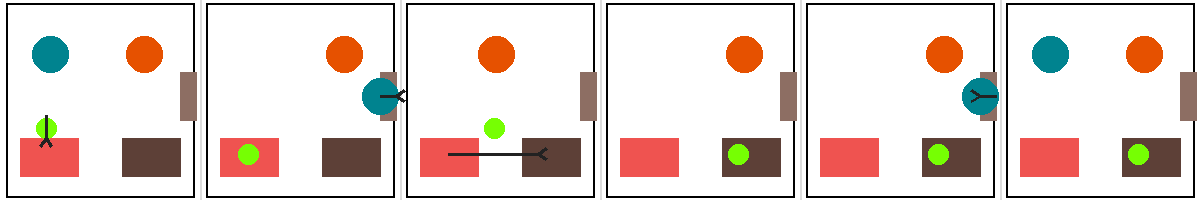}{sally\_anne\_search} &
\gridpanel{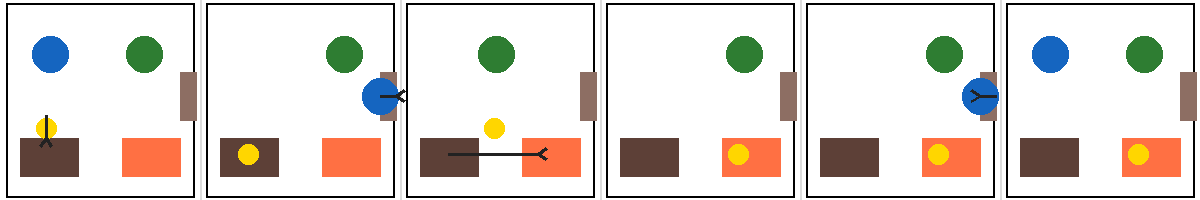}{sally\_anne\_belief} \\[6pt]
\gridpanel{s3_sally_anne_reality_000.png}{sally\_anne\_reality} &
\gridpanel{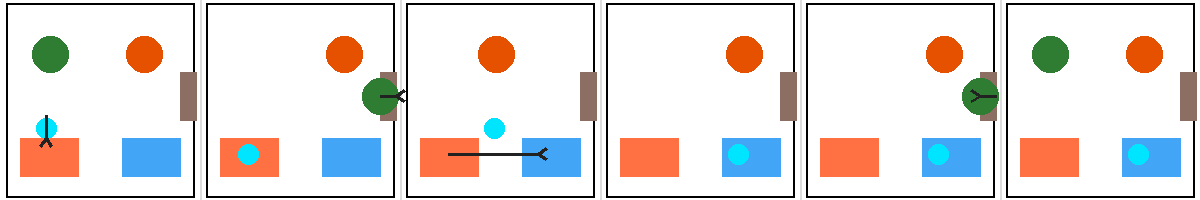}{reality\_vs\_belief} \\[6pt]
\gridpanel{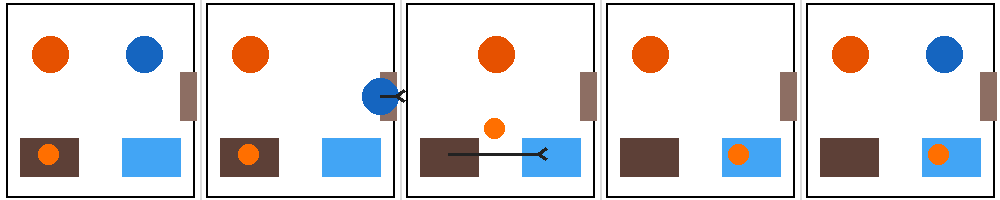}{knowledge\_who\_knows} &
\gridpanel{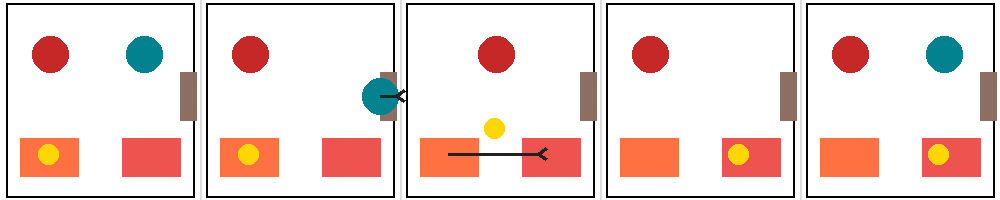}{knowledge\_belief} \\[6pt]
\gridpanel{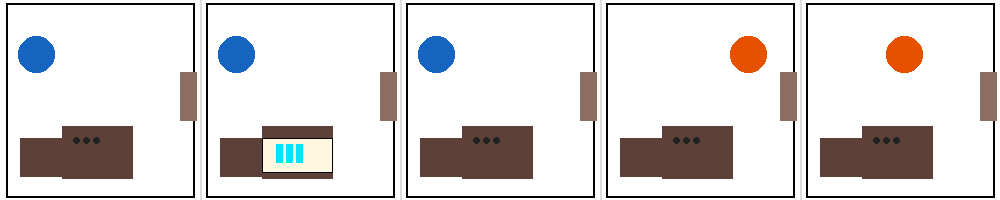}{unexpected\_contents} &
\gridpanel{s3_two_rooms_000.png}{two\_rooms} \\[6pt]
\gridpanel{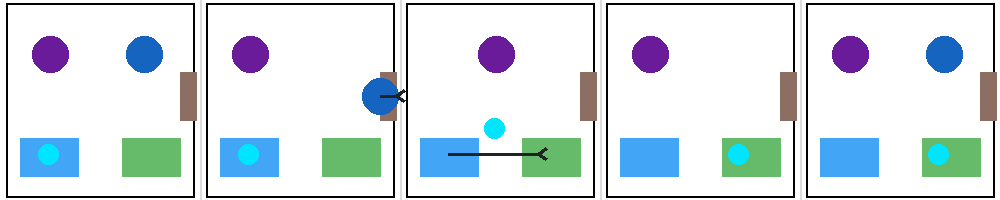}{perspectives\_differ} &
\gridpanel{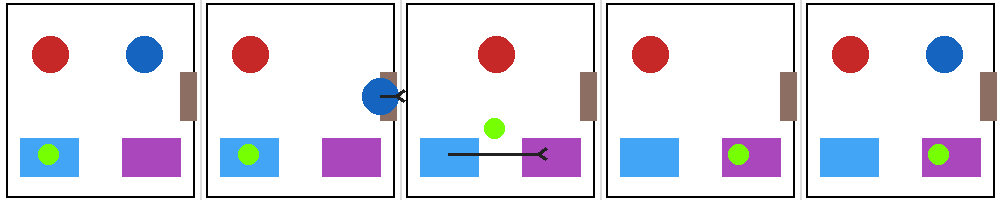}{second\_order} \\
\end{tabular}}
\end{center}

\subsection*{3. Representative panel example for each template}

\textbf{Example outputs.} The cases below are sample examples from the saved
probes. They use specific saved probe images and report saved responses from
\texttt{gpt-4o} for conditions \texttt{A}, \texttt{B\_p},
\texttt{B\_f}, and \texttt{C}. The percentages reported below are
template-level averages across 350 trials (7 models $\times$ 50 probes), not
just the single illustrated case.

\caseheader
{sally\_anne\_reality}
{3}
{s3_sally_anne_reality_000.png}
{purple and orange are in a room with a left mint container and a right violet container. purple places the pink object in the left container and then leaves. While purple is away, orange moves the object to the right container. purple returns after the move.}

\textbf{Why this case is interesting.} This is a clean reality-location
control where the image-only condition can still fail even though the answer
depends only on the object's true final location. In this saved example,
\texttt{A} fails while \texttt{B\_p}, \texttt{B\_f}, and
\texttt{C} all succeed. Calculated over the full template-level evaluation
set, the condition averages are: \texttt{A = 74.9\%},
\texttt{B\_p = 98.9\%},
\texttt{B\_f = 99.4\%}, \texttt{C = 100.0\%}.

\textbf{Model outputs by condition}
\begin{center}
\footnotesize
\setlength{\tabcolsep}{4pt}
\renewcommand{\arraystretch}{1.08}
\begin{tabular}{@{}L{0.10\linewidth}L{0.18\linewidth}L{0.50\linewidth}L{0.16\linewidth}@{}}
\toprule
\textbf{Variant} & \textbf{Setup} & \textbf{Prompt used for model} & \textbf{Model output} \\
\midrule
\outputrow{A}{Image only}{Study this 6-panel visual sequence. Where is the pink object actually located in the final panel, left or right? Answer ONLY 'left' or 'right'.}{left}
\outputrow{B\_p}{Image + perceptual scaffold}{Study this image. Here is a structured perceptual description: Panel 1: purple and orange characters are in the room. pink object is near the left mint container. purple character places the object there. Panel 2: purple character moves to the door and exits. orange character remains in the room. pink object is in the left mint container. Panel 3: Only orange character is in the room. pink object moves from the left mint container toward the right violet container. Panel 4: orange character is in the room. pink object is in the right violet container. Panel 5: purple character returns through the door. orange character is in the room. pink object is in the right violet container. Panel 6: purple and orange characters are both in the room. pink object is in the right violet container. Where is the pink object actually located in the final panel, left or right? Answer ONLY 'left' or 'right'.}{right}
\outputrow{B\_f}{Image + full scaffold}{Study this image. Here is a full structured description: Panel 1: purple and orange characters are in the room. pink object is near the left mint container. purple character places the object there. Panel 2: purple character moves to the door and exits. orange character remains in the room. pink object is in the left mint container. Panel 3: Only orange character is in the room. pink object moves from the left mint container toward the right violet container. Panel 4: orange character is in the room. pink object is in the right violet container. Panel 5: purple character returns through the door. orange character is in the room. pink object is in the right violet container. purple character was outside when the object moved from left to right and did not see that move. Where is the pink object actually located in the final panel, left or right? Answer ONLY 'left' or 'right'.}{right}
\outputrow{C}{Text only}{purple and orange are in a room with a left mint container and a right violet container. purple places the pink object in the left container and then leaves. While purple is away, orange moves the object to the right container. purple returns after the move. Where is the pink object actually located in the final panel, left or right? Answer ONLY 'left' or 'right'.}{right}
\bottomrule
\end{tabular}
\end{center}

\caseheader
{two\_rooms}
{3}
{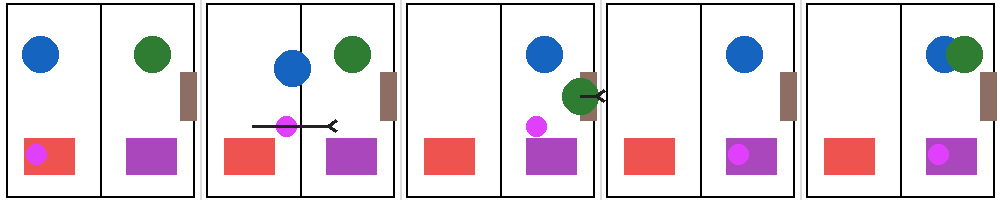}
{blue starts in the left room with the pink object. green is in the right room. blue carries the object into the right room while green is watching. green then leaves. While green is gone, blue places the object in the right violet container. green returns later.}

\textbf{Why this case is interesting.} This is a strong single-token failure
case: \texttt{A = 56.0\%}, \texttt{B\_p = 65.7\%},
\texttt{B\_f = 98.9\%}, \texttt{C = 78.6\%}. In the saved example below,
\texttt{A} and \texttt{B\_p} fail, while \texttt{B\_f} and
\texttt{C} both succeed.

\textbf{Model outputs by condition}
\begin{center}
\footnotesize
\setlength{\tabcolsep}{4pt}
\renewcommand{\arraystretch}{1.08}
\begin{tabular}{@{}L{0.10\linewidth}L{0.18\linewidth}L{0.50\linewidth}L{0.16\linewidth}@{}}
\toprule
\textbf{Variant} & \textbf{Setup} & \textbf{Prompt used for model} & \textbf{Model output} \\
\midrule
\outputrow{A}{Image only}{Study this 5-panel visual sequence. Where does the green character think the pink object is, left or right? Answer ONLY 'left' or 'right'.}{left}
\outputrow{B\_p}{Image + perceptual scaffold}{Study this image. Here is a structured perceptual description: Panel 1: blue character is in the left room with the pink object near the left coral container. green character is in the right room. Panel 2: blue character carries the pink object from the left room toward the right room. green character can see this transfer. Panel 3: green character exits from the right room. blue character and the pink object are in the right room. Panel 4: Only blue character is present. pink object is placed in the right violet container. Panel 5: green character returns to the right room. pink object is in the right violet container. Where does the green character think the pink object is, left or right? Answer ONLY 'left' or 'right'.}{left}
\outputrow{B\_f}{Image + full scaffold}{Study this image. Here is a full structured description: Panel 1: blue character is in the left room with the pink object near the left coral container. green character is in the right room. Panel 2: blue character carries the pink object from the left room toward the right room. green character can see this transfer. Panel 3: green character exits from the right room. blue character and the pink object are in the right room. Panel 4: Only blue character is present. pink object is placed in the right violet container. Panel 5: green character returns to the right room. pink object is in the right violet container. green character left before the object was placed in the right container, but had already seen the object taken into the right room. Where does the green character think the pink object is, left or right? Answer ONLY 'left' or 'right'.}{right}
\outputrow{C}{Text only}{blue starts in the left room with the pink object. green is in the right room. blue carries the object into the right room while green is watching. green then leaves. While green is gone, blue places the object in the right violet container. green returns later. Where does the green character think the pink object is, left or right? Answer ONLY 'left' or 'right'.}{right}
\bottomrule
\end{tabular}
\end{center}

\caseheader
{unexpected\_contents}
{3}
{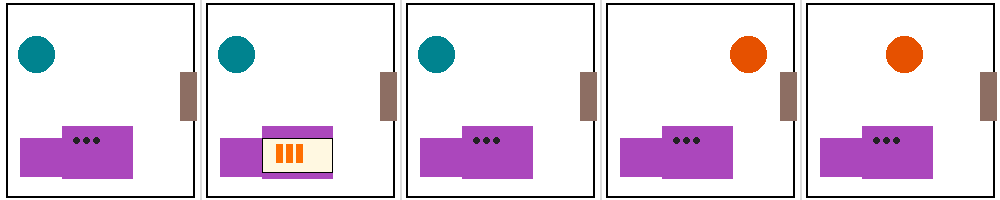}
{A violet box has round symbols on it. teal opens it and finds amber rectangular contents inside. The box is closed again. Later, orange arrives and sees only the closed box with the round symbols.}

\textbf{Why this case is interesting.} This is one of the two multi-token
templates in the benchmark. In this saved example, GPT-4o fails in
\texttt{A} but succeeds in \texttt{B\_p}, \texttt{B\_f}, and
\texttt{C}, which makes the image-only gap directly visible. Calculated over
the full template-level evaluation set, the condition averages are:
\texttt{A = 55.4\%}, \texttt{B\_p = 100.0\%},
\texttt{B\_f = 99.7\%}, \texttt{C = 100.0\%}.

\textbf{Model outputs by condition}
\begin{center}
\footnotesize
\setlength{\tabcolsep}{4pt}
\renewcommand{\arraystretch}{1.08}
\begin{tabular}{@{}L{0.10\linewidth}L{0.18\linewidth}L{0.50\linewidth}L{0.16\linewidth}@{}}
\toprule
\textbf{Variant} & \textbf{Setup} & \textbf{Prompt used for model} & \textbf{Model output} \\
\midrule
\outputrow{A}{Image only}{Study this 5-panel visual sequence. What does the orange character think is inside the box, round objects or yellow rectangles? Answer ONLY 'round objects' or 'yellow rectangles'.}{yellow rectangles}
\outputrow{B\_p}{Image + perceptual scaffold}{Study this image. Here is a structured perceptual description: Panel 1: teal character sees a closed violet box marked with round symbols. Panel 2: teal character opens the violet box. Inside are amber rectangular contents, not round objects. Panel 3: The violet box is closed again and still shows the round symbols. Panel 4: orange character arrives and sees only the closed violet box with round symbols. Panel 5: orange character is still looking at the closed violet box. What does the orange character think is inside the box, round objects or yellow rectangles? Answer ONLY 'round objects' or 'yellow rectangles'.}{round objects}
\outputrow{B\_f}{Image + full scaffold}{Study this image. Here is a full structured description: Panel 1: teal character sees a closed violet box marked with round symbols. Panel 2: teal character opens the violet box. Inside are amber rectangular contents, not round objects. Panel 3: The violet box is closed again and still shows the round symbols. Panel 4: orange character arrives and sees only the closed violet box with round symbols. Panel 5: orange character is still looking at the closed violet box. orange character never saw the box opened and only saw the closed box with the round symbols. What does the orange character think is inside the box, round objects or yellow rectangles? Answer ONLY 'round objects' or 'yellow rectangles'.}{round objects}
\outputrow{C}{Text only}{A violet box has round symbols on it. teal opens it and finds amber rectangular contents inside. The box is closed again. Later, orange arrives and sees only the closed box with the round symbols. What does the orange character think is inside the box, round objects or yellow rectangles? Answer ONLY 'round objects' or 'yellow rectangles'.}{round objects}
\bottomrule
\end{tabular}
\end{center}

\caseheader
{second\_order}
{3}
{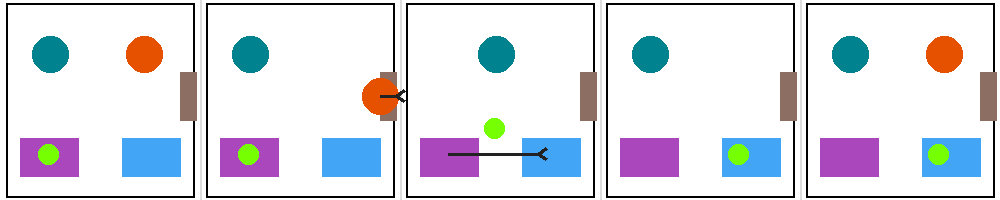}
{teal and orange both see the lime object in the left violet container. orange leaves. While orange is away, teal moves the object to the right light-blue container. orange returns after the move.}

\textbf{Why this case is interesting.} This is a higher-order belief probe:
the model must judge whether one agent knows that another agent now has a
false belief. In this saved example, \texttt{A} fails while
\texttt{B\_p}, \texttt{B\_f}, and \texttt{C} succeed. Calculated
over the full template-level evaluation set, the condition averages are:
\texttt{A = 64.9\%}, \texttt{B\_p = 97.7\%},
\texttt{B\_f = 100.0\%}, \texttt{C = 90.9\%}.

\textbf{Model outputs by condition}
\begin{center}
\footnotesize
\setlength{\tabcolsep}{4pt}
\renewcommand{\arraystretch}{1.08}
\begin{tabular}{@{}L{0.10\linewidth}L{0.18\linewidth}L{0.50\linewidth}L{0.16\linewidth}@{}}
\toprule
\textbf{Variant} & \textbf{Setup} & \textbf{Prompt used for model} & \textbf{Model output} \\
\midrule
\outputrow{A}{Image only}{Study this 5-panel visual sequence. Does the teal character know that the orange character has a wrong belief about the object's location, yes or no? Answer ONLY 'yes' or 'no'.}{no}
\outputrow{B\_p}{Image + perceptual scaffold}{Study this image. Here is a structured perceptual description: Panel 1: teal and orange characters are in the room. lime object is in the left violet container. Panel 2: orange character exits through the door. teal character remains. lime object is in the left violet container. Panel 3: Only teal character is in the room. lime object moves from the left violet container toward the right light-blue container. Panel 4: teal character is in the room. lime object is in the right light-blue container. Panel 5: orange character returns. teal character is in the room. lime object is in the right light-blue container. Does the teal character know that the orange character has a wrong belief about the object's location, yes or no? Answer ONLY 'yes' or 'no'.}{Yes.}
\outputrow{B\_f}{Image + full scaffold}{Study this image. Here is a full structured description: Panel 1: teal and orange characters are in the room. lime object is in the left violet container. Panel 2: orange character exits through the door. teal character remains. lime object is in the left violet container. Panel 3: Only teal character is in the room. lime object moves from the left violet container toward the right light-blue container. Panel 4: teal character is in the room. lime object is in the right light-blue container. Panel 5: orange character returns. teal character is in the room. lime object is in the right light-blue container. orange character did not see the move, so orange still has the old belief. teal knows the move happened and knows that orange missed it. Does the teal character know that the orange character has a wrong belief about the object's location, yes or no? Answer ONLY 'yes' or 'no'.}{Yes.}
\outputrow{C}{Text only}{teal and orange both see the lime object in the left violet container. orange leaves. While orange is away, teal moves the object to the right light-blue container. orange returns after the move. Does the teal character know that the orange character has a wrong belief about the object's location, yes or no? Answer ONLY 'yes' or 'no'.}{yes}
\bottomrule
\end{tabular}
\end{center}

\subsection*{4. Failures at the text-only ceiling}

The text-only ceiling is high ($94.5\%$ aggregate) but not perfect, and its failures are informative. On \textsc{two\_rooms}, text-only accuracy is $78.6\%$. In the saved case below (Figure~\ref{fig:qualitative}b; probe \texttt{s3\_two\_rooms\_000}; expected answer ``right''), GPT-4o (2024-11-20) receives the full ground-truth narration, in which the returning agent watched the object being carried into the right room before leaving, yet answers ``left''. The model over-applies the heuristic ``absent agent, therefore old location'' instead of tracking what the agent actually witnessed. A second text-only failure family is the Claude Sonnet~4 Stage~2 direction-convention anomaly (\S\ref{app:addlimitations}; $51.5\%$). These cases show that residual inferential errors exist in text as well; our claim is comparative, concerning the $20.6$ pp visual-to-text gap, not a claim that text-side reasoning is perfect.

\begin{center}
\footnotesize
\setlength{\tabcolsep}{4pt}
\renewcommand{\arraystretch}{1.08}
\begin{tabular}{@{}L{0.10\linewidth}L{0.18\linewidth}L{0.50\linewidth}L{0.16\linewidth}@{}}
\toprule
\textbf{Variant} & \textbf{Setup} & \textbf{Prompt used for model} & \textbf{Model output} \\
\midrule
C & Text only & red starts in the left room with the cyan object. blue is in the right room. red carries the object into the right room while blue is watching. blue then leaves. While blue is gone, red places the object in the right brown container. blue returns later. Where does the blue character think the cyan object is, left or right? Answer ONLY 'left' or 'right'. & left \\
\bottomrule
\end{tabular}
\end{center}



\end{document}